\documentclass{article}

\usepackage[final]{neurips_2024}

\usepackage[utf8]{inputenc} 
\usepackage[T1]{fontenc}    
\usepackage{hyperref}       
\usepackage{url}            
\usepackage{booktabs}       
\usepackage{amsfonts}       
\usepackage{nicefrac}       
\usepackage{microtype}      
\usepackage{xcolor}         

\title{Generative Adversarial Model-Based Optimization via Source Critic Regularization}

\author{%
  Michael S.~Yao\thanks{Correspondence to: \texttt{michael.yao@pennmedicine.upenn.edu}, \texttt{obastani@seas.upenn.edu}} \\
  Department of Bioengineering\\
  Perelman School of Medicine\\
  University of Pennsylvania \\
  \And
  Yimeng~Zeng \\
  Department of Computer \\
  and Information Science \\
  University of Pennsylvania\\
  \And
  Hamsa~Bastani \\
  The Wharton School\\
  University of Pennsylvania\\
  \And
  Jacob~Gardner \\
  Department of Computer \\
  and Information Science \\
  University of Pennsylvania\\
  \And
  James C.~Gee \\
  Department of Radiology \\
  University of Pennsylvania \\
  \And
  Osbert~Bastani$^*$ \\
  Department of Computer \\
  and Information Science \\
  University of Pennsylvania\\
}

\begin{document}

\maketitle

\begin{abstract}
Offline model-based optimization seeks to optimize against a learned surrogate model without querying the true oracle objective function during optimization. Such tasks are commonly encountered in protein design, robotics, and clinical medicine where evaluating the oracle function is prohibitively expensive. However, inaccurate surrogate model predictions are frequently encountered along offline optimization trajectories. To address this limitation, we propose \textit{generative adversarial model-based optimization} using \textbf{adaptive source critic regularization (aSCR)}\textemdash a task- and optimizer- agnostic framework for constraining the optimization trajectory to regions of the design space where the surrogate function is reliable. We propose a computationally tractable algorithm to dynamically adjust the strength of this constraint, and show how leveraging aSCR with standard Bayesian optimization outperforms existing methods on a suite of offline generative design tasks. Our code is available at \href{https://github.com/michael-s-yao/gabo}{https://github.com/michael-s-yao/gabo}.
\end{abstract}

\section{Introduction}

In many real-world tasks, we often seek to optimize the value of an objective function over some search space of inputs. Such optimization problems span across a wide variety of domains, including molecule and protein design \citep{organic, guacamol, maus2022}, patient treatment effect estimation \citep{kim2021, ite2022, xu2022}, and resource allocation in public policy \citep{covid2021, sex2021}. A number of algorithms have been explored for online optimization in these domains, including first-order methods, quasi-Newton methods,
and Bayesian optimization \citep{8903465}.

However, in many situations it may prove difficult or costly to estimate the objective function for any arbitrary input configuration. Evaluating newly proposed molecules requires expensive experimental laboratory setups, and testing multiple drug doses for a single patient can potentially be dangerous. In these scenarios, the allowable budget for objective function queries is prohibitive, thereby limiting the utility of out-of-the-box online policy optimization methods.

\begin{wrapfigure}{r}{0.5\textwidth}
\vspace{-2ex}
\begin{center}
\centerline{\includegraphics[width=0.5\textwidth]{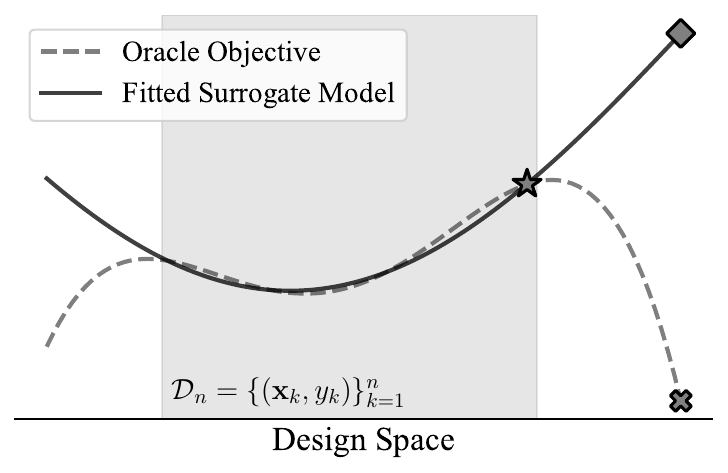}}
\caption{Na\"{i}ve offline model-based optimization (MBO) \citep{coms}, which optimizes against a learned surrogate model $f_{\theta}$ trained on a fixed dataset $\mathcal{D}_n=\{(\mathbf{x}_i, y_i)\}_{i=1}^n$ (shaded region) without access to the true oracle $f$, often yields candidate designs $\mathbf{x}^*$ (i.e., diamond) that score poorly using the true oracle (i.e., cross). Our method (aSCR) constrains optimization trajectories to avoid these extrapolated points, instead proposing `in-distribution' designs (i.e., star).}
\label{fig:overview}
\end{center}
\vspace{-6ex}
\end{wrapfigure}

To overcome this limitation, recent work has investigated the utility of optimization methods in the \textit{offline} setting, where we are unable to query the objective function during the optimization process and instead only have access to a set of prior observations of inputs and associated objective values; this problem can often be referred to as offline \textit{model-based optimization} (\textbf{MBO}) \citep{coms, bonet}. While one may na\"{i}vely attempt to learn a surrogate black-box model from the prior observations that approximates the true oracle objective function, such models can suffer from overestimation errors, yielding falsely promising objective estimates for inputs not contained in the offline dataset. As a result, offline optimization against the surrogate objective may yield low-scoring candidate designs according to the true oracle objective function\textemdash a key limitation of traditional policy optimization techniques in the offline setting (\textbf{Fig. \ref{fig:overview}}).

In this work, we propose a novel offline MBO algorithm that leverages source critic models to optimize a surrogate objective while simultaneously remaining in-distribution when compared against a reference offline dataset. In this setting, an optimizer is rewarded for proposing optima that are ``similar'' to reference data points, thereby minimizing overestimation error and allowing for more robust oracle function optimization in the offline setting. Inspired by recent work on generative adversarial networks \citep{gan}, we quantify design similarity by proposing a novel method that regularizes a surrogate objective model using a source critic actor, which we call \textit{adaptive source critic regularization} (\textbf{aSCR}). We show how our algorithm can be readily leveraged with optimization methods such as Bayesian optimization (BO) and first-order methods.

\textbf{Contributions:} Our contributions are as follows: (1) We propose a novel approach for MBO that formulates the task as a constrained primal optimization problem, and we show how this framework can be used to solve for the optimal tradeoff between na\"{i}vely optimizing against the surrogate model and staying in-distribution relative to the offline dataset. (2) We introduce a computationally tractable method\textemdash which we call adaptive source critic regularization (aSCR)\textemdash to implement this framework with two popular optimization methods: Bayesian optimization and gradient ascent. (3) We show that compared to prior methods, our proposed algorithm with Bayesian optimization empirically achieves the highest rank of \textbf{3.8} (second best is 5.5) on top-1 design evaluation, and highest rank of \textbf{3.0} (second best is 4.6) on top-128 design evaluation across a variety of tasks spanning multiple scientific domains.

\section{Related Work}

Leveraging source critic model feedback for adversarial training of neural networks was popularized by works such as \citet{gan}, where a generator and adversarial discriminator ``play'' a zero-sum minimax game to train a generative model. However, such discriminators often suffer from mode collapse and training instability in practice. To overcome these limitations, \citet{wgan} introduced the Wasserstein generative adversarial net (WGAN), which instead utilizes a source critic that learns to approximate the 1-Wasserstein distance between the generated and training distributions. However, WGANs and similar networks primarily aim to generate samples that look in-distribution from a latent space prior, rather than optimize against an objective function. In our work, we adapt WGAN-inspired source critic models for Wasserstein distance estimation.

Separately in the field of optimization,
\citet{cbas} introduced a method for conditioning by adaptive sampling (CbAS) that learns a density model of the input space that is gradually adapted towards the optimal solution. However, such prior works have focused on solving low-dimensional, online optimization tasks \citep{cmaes, cbas}. More recently, \citet{coms} introduced conservative objective models (\textbf{COM}) specifically for \textit{offline} optimization tasks; however, their method requires directly modifying the parameters of the surrogate function during optimization, which is not always feasible for any general task. \citet{bonet} proposed Black-box Optimization Networks (\textbf{BONET}) to learn the dynamics of optimization trajectories using a causally masked transformer model, and \citet{ddom} introduced Denoising Diffusion Optimization Models (\textbf{DDOM}) to learn the generative process via a diffusion model. Furthermore, \citet{roma} and \citet{bdi} describe Robust Model Adaptation (\textbf{RoMA}) and Bidirectional learning via infinite-width networks (\textbf{BDI}), respectively. RoMA regularizes the gradient of surrogate objective models by enforcing a local smoothness prior at the observed inputs, and BDI learns bidirectional mappings between low- and high- scoring candidates. Finally, \citet{expt} introduce Experiment Pretrained Transformers (\textbf{ExPT}) to learn a general model for optimization using unsupervised methods. While these recent works and others propose promising algorithms for offline optimization tasks, they are often evaluated using expensive oracle query budgets that are often not achievable in practice\textemdash especially for potentially dangerous tasks such as patient care and other high-stakes applications.

\section{Background}

\subsection{Offline Model-Based Optimization}
In many real-world domains, we often seek to optimize an \textit{oracle} objective function $f(\mathbf{x})$ over a space of design candidates $\mathcal{X}$ to solve for $\mathbf{x}^*=\text{argmax}_{\mathbf{x}\in\mathcal{X}} f(\mathbf{x})$. Examples of such problems include optimizing certain desirable properties of molecules in molecular design \citep{organic, guacamol, maus2022}, and estimating the optimal therapeutic intervention for patient care in clinical medicine \citep{kim2021, ite2022, xu2022}. In practice, however, the true objective function $f$ may be costly to compute or even entirely unknown, making it difficult to query in optimizing $f(\mathbf{x})$. Instead, it is often more feasible to obtain access to a reference labeled dataset of observations from nature $\mathcal{D}_n=\{(\mathbf{x}_1, y_1), \ldots, (\mathbf{x}_n, y_n)\}$ where $y_i=f(\mathbf{x}_i)$. Optimization methods may use a variety of different strategies to leverage $\mathcal{D}_n$ in the offline setting \citep{bonet, ddom, bdi}; one common approach used by \citet{coms} and others is to learn a regressor model $f_\theta$ parametrized by
\begin{equation}
\theta^*=\text{argmin}_{\theta}\enspace \mathbb{E}_{(\mathbf{x}_i, y_i)\sim\mathcal{D}_n} ||f_{\theta}(\mathbf{x}_i)-y_i||^2
\end{equation}
as a \textit{surrogate model} for the true oracle objective $f(\mathbf{x})$. Rather than querying the oracle $f$ as in the online setting, we can instead solve the related optimization problem
\begin{equation}
\mathbf{x}^*=\text{argmax}_{\mathbf{x}\in\mathcal{X}} f_{\theta}(\mathbf{x}) \label{eq:mbo}
\end{equation}
with the hope that optimizing $f_{\theta}$ will also lead to desirable oracle values of $f$ as well. Solving (\ref{eq:mbo}) is one instantiation of offline \textbf{model-based optimization (MBO)} for which a number of techniques have been developed, such as gradient ascent and Bayesian optimization (\textbf{BO}) \citep{8903465}.

Of note, it is difficult to guarantee the reliability of the model's predictions for $\mathbf{x}\notin\mathcal{D}_n$ that are almost certainly encountered in the optimization trajectory. Thus, na\"{i}vely optimizing the surrogate objective $f_{\theta}$ can result in ``optima'' that are low-scoring  according to the oracle objective $f$.

\subsection{Optimization Over Latent Spaces}

In certain cases, the search space $\mathcal{X}$ for an optimization task may be discretized over a finite set of structured inputs, such as amino acids for protein sequences or atomic building blocks for molecules. However, many historical optimization algorithms do not generalize well to these settings for a number of different reasons, such as the lack of gradients with respect to the input designs to guide the optimization trajectory. Instead of directly optimizing over $\mathcal{X}$, recent work leverages deep variational autoencoders (VAEs) to first map the input space into a continuous, (often) lower dimensional latent space $\mathcal{Z}$ and then performing optimization over $\mathcal{Z}$ instead \citep{latentopt1_2020, latentopt2_2021, maus2022}. A VAE is composed of a two components: (1) an encoder with parameters $\phi$ that learns an approximated posterior distribution $q_{\phi}(z | \mathbf{x})$ for $\mathbf{x}\in\mathcal{X}, z\in\mathcal{Z}$; and (2) a decoder with parameters $\varphi$ that learns the conditional likelihood distribution $p_{\varphi}(\mathbf{x} | z)$ \citep{vae}. The encoder and decoder are co-trained to maximize the evidence lower bound (ELBO)
\begin{equation}
\text{ELBO}=\mathbb{E}_{z\sim q_{\phi}}\left[\log p_{\varphi}(\mathbf{x} | z)\right]-D_{\text{KL}}\left[q_{\phi}(z | \mathbf{x})\text{ }||\text{ }p_{\text{VAE}}(z)\right] \label{eq:elbo}
\end{equation}
where $D_{\text{KL}}$ is the Kullback-Leibler (KL) divergence and $p_{\text{VAE}}(z)$ is the prior distribution. A common choice is to set  $p_{\text{VAE}}=\mathcal{N}(0, I)$ (i.e., the standard normal distribution). Optimization can then be performed over the continuous \textit{latent space} $\mathcal{Z}$ of the VAE to propose `latent space designs' that can be readily decoded using the decoder $\varphi$ back into the original input space.

One such optimization method over VAE latent spaces is \textbf{Bayesian optimization (BO)}, a sample-efficient framework for solving expensive black-box optimization problems \citep{mockus1982, osborne2009, snoek2012}. While the utility of BO has primarily been explored for expensive-to-evaluate black-box functions in prior literature, recent work has shown that BO also outperforms baseline optimization methods in offline tasks involving models that are relatively inexpensive to evaluate, such as the neural network surrogates used in model-based optimization (MBO). Multiple prior works have shown that BO and related methods consistently outperform both first-order gradient-based and stochastic evolutionary methods \citep{eriksson2019, maus2022, hvarfner2024, eriksson2021, astudillo2019}.

\subsection{Wasserstein Metric Between Probability Distributions}
The Wasserstein distance is a distance metric between any two probability distributions, and is closely related to problems in optimal transport. We define the $p=1$ Wasserstein distance between a reference distribution $P$ and a generated distribution $Q$ using distance metric $d(\cdot, \cdot)$ as
\begin{equation}
W_1(P, Q)=\inf_{\gamma\in\Gamma(P, Q)}\mathbb{E}_{(z', z)\sim \gamma} d(z', z) \label{eq:wasserstein}
\end{equation}
where $\Gamma$ is the set of all couplings between $P$ and $Q$. For empirical distributions where $p_n$ ($q_n$) is based on $n$ observations $\{z_j'\}_{j=1}^n$ ($\{z_i\}_{i=1}^n$), (\ref{eq:wasserstein}) can be simplified to
\begin{equation}
W_1(p_n, q_n)=\inf_{\sigma}\frac{1}{n}\sum_{i=1}^n||z'_{\sigma(i)}-z_i|| \label{eq:empirical-wasserstein}
\end{equation}
where the infimum is over all permutations $\sigma$ of $n$ elements. Leveraging the Kantorovich-Rubinstein duality theorem \citep{wasserstein-duality}, (\ref{eq:empirical-wasserstein}) can be equivalently written as
\begin{equation}
W_1(p_n, q_n)=\frac{1}{K}\sup_{||c||_L\leq K} \Big[\mathbb{E}_{z'\sim P}[c(z')]-\mathbb{E}_{z\sim Q}[c(z)]\Big] \label{eq:kr-duality}
\end{equation}
where $c(z)$ is a \textit{source critic} and $||c||_L$ is the Lipschitz norm of $c(z)$. In the Wasserstein GAN (WGAN) model proposed by \citet{wgan}, a generative network and source critic are co-trained in a minimax game where the generator (critic) seeks to minimize (maximize) the Wasserstein distance $W_1$ between the training and generated distributions. Such an optimization schema enables the generator policy to learn the distribution of training samples from nature.

\section{A Framework for Generative Adversarial Optimization}

In this section we describe our proposed framework for generative adversarial model-based optimization using \textbf{adaptive source critic regularization} (\textbf{aSCR}). Our method uses a $K$-Lipschitz source critic model to dynamically regularize the optimization objective to avoid extrapolation against the proxy surrogate model $f_\theta$ in offline MBO.

\subsection{Constrained Optimization Formulation}

In offline generative optimization, we aim to optimize against a surrogate objective function $f_{\theta}$. In order to ensure that we are achieving reliable estimates of the true, unknown oracle objective, we can add a regularization penalty to keep generated samples ``similar'' to those from the training dataset of $f_{\theta}$ according to an adversarial source critic trained to discriminate between generated and offline samples. That is, in contrast to (\ref{eq:mbo}), aSCR instead considers a closely related \textit{constrained} problem
\begin{equation}
\begin{split}
\text{minimize}_{z\in\mathcal{Z}}&\quad -f_{\theta}(z)\\
\text{subject to}&\quad \mathbb{E}_{z'\in P} [c^*(z')] - c^*(z)\leq 0 \label{eq:constrained-opt}
\end{split}
\end{equation}
over some configuration space $\mathcal{Z}\subseteq\mathbb{R}^d$, and where we define $c^*$ as a source critic model that maximizes $\mathbb{E}_{z'\in P} [c^*(z')] - \mathbb{E}_{z\in Q}[c^*(z)]$ over all $K$-Lipschitz functions as in (\ref{eq:kr-duality}). We can think of $\mathbb{E}_{z'\in P} [c^*(z')] - c^*(z)$ as the contribution of a particular generated datum $z$ to the overall $p=1$ Wasserstein distance between the generated candidate ($Q$) and reference ($P$) distributions of designs as in (\ref{eq:kr-duality}). In practice, we model $c^*$ as a fully connected neural net. Intuitively, the imposed constraint restricts the feasible search space to designs that score at least as in-distribution as the average sample in the offline dataset according to the source critic. Therefore, $c^*$ acts as an adversarial model to regularize the optimization policy. Of note, this additional constraint in (\ref{eq:constrained-opt}) may be highly non-convex for general $c^*$, and so it is often impractical to directly apply (\ref{eq:constrained-opt}) to any arbitrary MBO policy.

\subsection{Dual Formulation}

To solve this implementation problem, we instead look to reformulate (\ref{eq:constrained-opt}) in its dual space by first considering the Lagrangian $\mathcal{L}$ of our constrained problem:
\begin{equation}
\mathcal{L}\left(z; \lambda\right)=-f_{\theta}(z)+\lambda\left[\mathbb{E}_{z'\in P} [c^*(z')] - c^*(z)\right] \label{eq:lagrangian}
\end{equation}
where $\lambda\geq 0$ is the Lagrange multiplier associated with the constraint in (\ref{eq:constrained-opt}). We can equivalently think of $\lambda$ as a hyperparameter that controls the relative strength of the source critic-penalty term: $\lambda=0$ equates to na\"{i}vely optimizing the surrogate objective, while $\lambda\gg 1$ asymptotically approaches a WGAN-like optimization policy. Minimizing $\mathcal{L}$ thus minimizes a relative sum of $-f_{\theta}$ and the Wasserstein distance contribution from any particular generated datum $z$ with relative weighting dictated by the hyperparameter $\lambda$. From duality, minimizing $\mathcal{L}$ over $z$ and simultaneously maximizing over $\lambda\in\mathbb{R}_+$ is equivalent to the original constrained problem in (\ref{eq:constrained-opt}).

The challenge now is in determining this optimal value of $\lambda$: if $\lambda$ is too small, then the objective estimates may be unreliable; if $\lambda$ is too large, then the optimization trajectory may be unable to adequately explore the input space. Prior work by \citet{coms} has previously explored the idea of formulating offline optimization problems as a similarly regularized Lagrangian (albeit with a separate regularization constraint), although their method tunes a static hyperparameter by hand. In contrast, aSCR treats $\lambda$ as a dynamic parameter that adapts to the optimization trajectory in real time.

\subsection{Computing the Lagrange Multiplier \texorpdfstring{$\lambda$}{Lambda}}

Continuing with our dual formulation of (\ref{eq:constrained-opt}), the Lagrange dual function $g(\lambda)$ is defined as $g(\lambda)=\inf_{z\in\mathbb{R}^n} \mathcal{L}\left(z; \lambda\right)$. The $z=\hat{z}$ that minimizes the Lagrangian in the definition of $g$ is evidently a function of $\lambda$. To show this, we use the first-order condition that $\nabla_z \mathcal{L}=0$ at $z=\hat{z}$. Per (\ref{eq:lagrangian}), we have
\begin{equation}
\nabla_{z}\mathcal{L}(\hat{z}; \lambda)=-\nabla_{z}f_{\theta}(\hat{z})-\lambda\nabla_zc^*(\hat{z})=0 \label{eq:6}
\end{equation}
In general, solving (\ref{eq:6}) for $\hat{z}$ is computationally intractable\textemdash especially in high-dimensional problems. Instead, we can approximate $\hat{z}$ by relaxing the condition in (\ref{eq:6}) according to
\begin{equation}
\begin{split}
\hat{z}(\lambda)=\mathop{\mathrm{argmin}}_{z\in\mathbb{R}^n}\frac{1}{2}\left|\left|-\nabla_{z}f_{\theta}(z)-\lambda\nabla_zc^*(z)\right|\right|^2 \label{eq:7}
\end{split}
\end{equation}
Our key insight is that although minimizing the loss term in (\ref{eq:7}) is not practical when the feasible set is na\"{i}vely uniform over $\mathbb{R}^n$, we can instead choose to focus our attention on latent space coordinates with high associated probability according to the VAE prior distribution $p_{\text{VAE}}(z)$. This is because in optimization problems acting over the latent space of any variational autoencoder, the majority of the encoded information content is embedded according to $p_{\text{VAE}}(z)$ due to the Kullback-Leibler (KL) divergence contribution to VAE training. Put simply, the encoder distribution $q_{\phi}(z | \mathbf{x})$ is trained so that $D_{\text{KL}}[q_{\phi}(z|\mathbf{x}) || p_{\text{VAE}}(z))]$ is optimized as a regularization term in (\ref{eq:elbo}). We argue that it is thus sufficient enough to approximate $\hat{z}(\lambda)$ using a Monte Carlo sampling schema with random samples $\mathcal{Z}_N=(z_1, z_2, \ldots, z_N)\sim p_{\text{VAE}}(z)$:
\begin{equation}
\begin{split}
\hat{z}(\lambda)\approx\mathop{\mathrm{argmin}}_{\mathcal{Z}_N\sim p_{\text{VAE}}(z)}\frac{1}{2}\left|\left|-\nabla_{z}f_{\theta}(z)-\lambda\nabla_zc^*(z)\right|\right|^2 \label{eq:8}
\end{split}
\end{equation}
We can now concretely write an approximation of the Lagrange dual problem of (\ref{eq:constrained-opt}):
\begin{equation}
\begin{split}
\text{maximize}&\quad g(\lambda)=-f_{\theta}(\hat{z})+\lambda\left[\mathbb{E}_{z'\in P} [c^*(z')] - c^*(\hat{z})\right]\\ 
\text{subject to}&\quad \lambda\geq 0 \label{eq:9}
\end{split}
\end{equation}
where $\hat{z}$ is as in (\ref{eq:8}). Defining the surrogate variable $\alpha$ such that $\lambda=\frac{\alpha}{1-\alpha}$, we can rewrite (\ref{eq:9}) as
\begin{equation}
\begin{split}
\text{maximize}&\quad -(1-\alpha)f_{\theta}(\hat{z})+\alpha\left[\mathbb{E}_{z'\in P} [c^*(z')] - c^*(\hat{z})\right]\\
\text{subject to}&\quad 0\leq\alpha< 1 \label{eq:10}
\end{split}
\end{equation}
In practice, we discretize the search space for $\alpha$ to 200 evenly spaced points between 0 and 1 inclusive. From weak duality, finding the optimal solution to (\ref{eq:9}) provides a lower bound on the optimal solution to the primal problem in (\ref{eq:constrained-opt}). \textbf{Algorithm \ref{alg:ascr}} can now be used to choose the optimal $\alpha$ (and hence $\lambda$) adaptively during offline optimization: we refer to our method as \textbf{Adaptive SCR (aSCR)}.

\subsection{Overall Algorithm}

Using Adaptive SCR, we now have a proposed method for dynamically computing $\alpha$ (and hence the Lagrange multiplier $\lambda$) of the constrained optimization problem in (\ref{eq:constrained-opt}). Importantly, aSCR can be integrated with any standard function optimization method by optimizing the Lagrangian objective in (\ref{eq:lagrangian}) over the candidate design space as opposed to the original unconstrained objective $f_{\theta}$. We refer to this algorithm as \textit{Generative Adversarial Model-Based Optimization} (GAMBO). To evaluate aSCR empirically, we instantiate two flavors of GAMBO: (1) \textbf{G}enerative \textbf{A}dversarial \textbf{B}ayesian \textbf{O}ptimization (\textbf{GABO}, \textbf{Algorithm \ref{alg:gabo}}); and (2) \textbf{G}enerative \textbf{A}dversarial \textbf{G}radient \textbf{A}scent (\textbf{GAGA}).\footnote{We detail the explicit algorithmic formulation for GAGA in \textbf{Supplementary Algorithm \ref{alg:gaga}}.}

We implement GABO using a quasi-expected improvement (qEI) acquisition function, iterative sampling budget of $T=32$, sampling batch size of $b=64$, and GAGA using a step size of $\eta=0.05$, $T=128$, and $b=16$. Of note, the optimization objective using aSCR is time-varying and causally linked to past observations made during the optimization process via intermittent training of the source critic $c$. Prior works from \citet{dynamic-bo} and \citet{dcbo} have examined optimization against dynamic objective functions, although have either entirely disregarded causal relationships between variables or only examined causality between inputs as opposed to inputs and the objective. We leave such methods for future work given that aSCR works well in practice.

\begin{wrapfigure}{R}{0.54\textwidth}
\vspace{-4ex}
\begin{minipage}{0.54\textwidth}
\begin{algorithm}[H]
  \caption{\small Adaptive Source Critic Regularization (SCR)}
\label{alg:ascr}
  \small\begin{algorithmic}
    \STATE {\bfseries Input:} differentiable surrogate objective $f_{\theta}: \mathbb{R}^d\rightarrow \mathbb{R}$, differentiable source critic $c: \mathbb{R}^d\rightarrow \mathbb{R}$, reference dataset $\mathcal{D}_n=\{z'_j\}_{j=1}^n$, $\alpha$ step size $\Delta\alpha$, search budget $\mathcal{B}$, norm threshold $\tau$
    \STATE Sample candidates $\mathcal{Z}_{\mathcal{B}}\leftarrow\{z_i\}_{i=1}^{\mathcal{B}}\sim \mathcal{N}(0, I_d)$
    \STATE Initialize $\alpha^*\leftarrow\text{None}$ and $g^*\leftarrow-\infty$
    \FOR{$\alpha$ {\bfseries in} $\text{range}(\text{start}=0, \text{end}=1, \text{stepsize}=\Delta\alpha)$}
      \STATE $z^*\leftarrow \text{argmin}_{z_i\in \mathcal{Z}_{\mathcal{B}}} ||(1-\alpha)\nabla f_{\theta}(z_i)+\alpha \nabla c(z_i)||_2$
      \IF{$||(1-\alpha)\nabla f_{\theta}(z^*)+\alpha \nabla c(z^*)||_2>\tau$}
        \STATE {\bfseries continue}  \hspace{5ex}// Discard $\alpha$ if best norm exceeds $\tau$
      \ENDIF
      \STATE $g\leftarrow -(1-\alpha)f_{\theta}(z^*)+\alpha\left[\mathbb{E}_{\mathcal{D}_n}[c(z'_j)] - c(z^*)\right]$
      \IF{$g>g^*$}
        \STATE $\alpha^*\leftarrow\alpha$ and $g^*\leftarrow g$ \hspace{9ex} // Implements (\ref{eq:10})
      \ENDIF
    \ENDFOR
    \STATE \textbf{return} $\alpha^*$
  \end{algorithmic}
\end{algorithm}
\vspace{-3ex}
\begin{algorithm}[H]
  \caption{\small Generative Adversarial BayesOpt (GABO)}
  \label{alg:gabo}
  \small\begin{algorithmic}
    \STATE {\bfseries Input:} surrogate objective $f_{\theta}: \mathbb{R}^d\rightarrow \mathbb{R}$, offline dataset $\mathcal{D}_n=\{z'_j\}_{j=1}^n$, acquisition function $a$, iterative sampling budget $T$, sampling batch size $b$, number of generator steps per source critic training $n_{\text{generator}}$, oracle query budget $k$
    \STATE {\bfseries AdaptiveSCR Input:} $\alpha$ step size $\Delta \alpha$, search budget $\mathcal{B}$, norm threshold $\tau$
    \STATE \textbf{Define:} Differentiable source critic $c: \mathbb{R}^d\rightarrow \mathbb{R}$
    \STATE {\bfseries Define:} Lagrangian $\mathcal{L}(z; \alpha):\mathbb{R}^d\times \mathbb{R}\rightarrow \mathbb{R}$ \hspace{3ex} // Eq. (\ref{eq:lagrangian})
    \STATE \hspace{6ex} $\mathcal{L}(z; \alpha)=-f_{\theta}(z)+\frac{\alpha}{1-\alpha}[\mathbb{E}_{z'\sim\mathcal{D}_n}[c(z')]-c(z)]$ 
    \STATE Sample candidates $\mathcal{Z}^1\leftarrow \{z^1_i\}_{i=1}^b \sim \text{SobolSequence}$
    \STATE // Train the source critic per Eq. (\ref{eq:kr-duality}) to optimality:
    \STATE $c \leftarrow \text{argmax}_{||c||_L\leq K} W_1(\mathcal{D}_n, \mathcal{Z}^1)$
    \STATE \hspace{2ex}$=\text{argmax}_{||c||_L\leq K}\left[\mathbb{E}_{z'\sim \mathcal{D}_n}[c(z')] - \mathbb{E}_{z\sim \mathcal{Z}^1}[c(z)]\right]$
    \STATE $\alpha\leftarrow \textbf{AdaptiveSCR}(f_{\theta}, c, \mathcal{D}_n, \Delta\alpha, \mathcal{B}, \tau)$ \hspace{4ex}// Alg. (\ref{alg:ascr})
    \STATE Evaluate candidates $\mathcal{Y}^1\leftarrow \{y^1_i\}_{i=1}^b=\{-\mathcal{L}(z^1_i; \alpha)\}_{i=1}^b$
    \STATE Place Gaussian Process (GP) prior on $f_{\theta}$
    \FOR{$t$ {\bfseries in} $2, 3, \ldots, T$}
      \STATE Update posterior on $f_{\theta}$ with $\mathcal{D}_{t-1}=\{(\mathcal{Z}^m, \mathcal{Y}^m)\}_{m=1}^{t-1}$
      \STATE Compute acquisition function $a$ using fitted posterior
      \STATE Sample candidates $\mathcal{Z}^t\leftarrow \{z_i^t\}_{i=1}^b$ according to $a$
      \STATE $\alpha\leftarrow \textbf{AdaptiveSCR}(f_{\theta}, c, \mathcal{D}_n, \Delta\alpha, \mathcal{B}, \tau)$
      \STATE Evaluate samples $\mathcal{Y}^t\leftarrow \{y^t_i\}_{i=1}^b=\{-\mathcal{L}(z_i^t; \alpha)\}_{i=1}^b$
      \IF{$t \text{ mod } n_{\text{generator}}$ equals $0$}
        \STATE // Train the source critic per Eq. (\ref{eq:kr-duality}) to optimality:
        \STATE $c \leftarrow \text{argmax}_{||c||_L\leq K} W_1(\mathcal{D}_n, \mathcal{Z}^t)$
        \STATE \hspace{2ex}$=\text{argmax}_{||c||_L\leq K}\left[\mathbb{E}_{z'\sim \mathcal{D}_n}[c(z')] - \mathbb{E}_{z\sim \mathcal{Z}^t}[c(z)]\right]$
      \ENDIF
    \ENDFOR
    \STATE \textbf{return} the top $k$ samples from the $T\times b$ observations
    \STATE \hspace{6ex} $\mathcal{D}_T=\{\{(z^m_i, y^m_i)\}_{i=1}^b\}_{m=1}^T$ according to $y^m_i$
  \end{algorithmic}
\end{algorithm}
\vspace{-8ex}
\end{minipage}
\end{wrapfigure}

\section{Experimental Evaluation}

\subsection{Datasets and Tasks}
\label{section:experimental}
To evaluate our proposed algorithm, we focus on a set of eight tasks spanning multiple domains with publicly available datasets in the field of offline model-based optimization. (1) The \textbf{Branin} function is a well-known synthetic benchmark function where the task is to maximize the two-dimensional Branin function $f_{br}: [-5, 10]\times[0, 15]\rightarrow \mathbb{R}$. (2) The \textbf{LogP} task is a well-studied optimization problem \citep{logp-ref1, logp-ref2, logp-ref3} where we search over candidate molecules to maximize the penalized water-octanol partition coefficient (logP) score, which is an approximate measure of a molecule's hydrophobicity \citep{logp} that also rewards structures that can be synthesized easily and feature minimal ring structures. We use the publicly available Guacamol benchmarking dataset from \citet{guacamol} to implement this task.

Tasks (3) - (7) are derived from Design-Bench, a publicly available set of MBO benchmarking tasks \citep{design-bench}: (3) \textbf{TF-Bind-8} aims to maximize the transcription factor binding efficiency of an 8-base-pair DNA sequence \citep{tfbind8}; (4) \textbf{GFP} the green fluorescence of a 237-amino-acid protein sequence \citep{cbas, gfp}; (5) \textbf{UTR} the gene expression from a 50-base-pair 5'UTR DNA sequence \citep{utr, utroracle}; (6) \textbf{ChEMBL} the mean corpuscular hemoglobin concentration (MCHC) biological response of a molecule using an offline dataset collected from the ChEMBL assay \texttt{CHEMBL3885882} \citep{chembl}; and (7) \textbf{D'Kitty} the morphological structure of the D'Kitty robot \citep{dkitty}.

Finally, (8) the \textbf{Warfarin} task uses the dataset of patients on warfarin medication from \citet{warfarin} to estimate the optimal dose of warfarin given clinical and pharmacogenetic patient data. Of note, in contrast to tasks (1) - (7) and other traditional MBO tasks in prior literature \citep{design-bench}, the Warfarin task is novel in that only a subset of the input design dimensions may be optimized over (i.e., warfarin dose) while the others remain fixed as conditioning variables (i.e., patient covariates). Such a task can therefore be thought of as \textit{conditional} model-based optimization.

\subsection{Policy Optimization and Evaluation}
\label{subsection:eval}
For all experiments, the surrogate objective model $f_{\theta}$ is a fully connected net with two hidden layers of size 2048 and LeakyReLU activations. $f_{\theta}$ takes as input a VAE-encoded latent space datum and returns the predicted objective function value as output. The VAE encoder and decoder backbone architectures vary by MBO task and are detailed in \textbf{Supplementary Table \ref{table:datasets}}. 
Following \citet{joint1} and \citet{maus2022}, we co-train the VAE and surrogate objective models together using an Adam optimizer 
with a learning rate of $3\times 10^{-4}$ for all tasks. For the optimization tasks over continuous design spaces (i.e., Branin, Warfarin, and D'Kitty), we fix the VAE encoder and decoders as the identity functions, such that the latent and input spaces are equivalent.

The source critic agent $c$ in (\ref{eq:constrained-opt}) is implemented as a fully connected net with two hidden layers with sizes equal to four (one) times the number of input dimensions for the first (second) layer. 
To constrain the Lipschitz norm of $c$ as in (\ref{eq:kr-duality}), we clamp the weights of the model between [-0.01, 0.01] after each optimization step as done by \citet{wgan}. The model is trained using gradient descent with a learning rate of 0.001 to maximize the Wasserstein distance between the dataset and generated candidates in the VAE latent space.

During optimization, both GABO and GAGA alternate between sampling new designs and training the source critic actor $c(z)$ until there is no improvement to the Wasserstein distance $W_1$ according to $c$ after 100 consecutive weight updates. We find that training $c$ every $n_{\text{generator}}=4$ sampling steps is a good choice across all tasks assessed, similar to prior work \cite{wgan}.

All MBO methods were evaluated using a fixed surrogate query budget of 2048. We focus on two evaluation metrics: 100th percentile (1) top $k=1$; and (2) top $k=128$ oracle score. The top $k=128$ evaluation metric is commonly reported in prior offline MBO literature \citep{bonet, coms, roma}; the top $k=1$ metric better accounts for the limited oracle query budget of the real-world tasks in which offline MBO would be of use. In both settings, an optimizer selects the top $k$ design that minimize the Lagrangian function value in (\ref{eq:lagrangian}) from the 2048 assessed designs to evaluate using the true oracle function, and the maximum score of those $k$ designs is reported across 10 random seeds.

We evaluate both GABO and GAGA against a number of pre-existing baseline algorithms on one internal cluster with 8 NVIDIA RTX A6000 GPUs. We include vanilla Bayesian Optimization (\textbf{BO}-qEI) and gradient ascent (\textbf{Grad.}) in our evaluation to assess the utility of our proposed aSCR algorithm. Furthermore, we evaluate limited-memory BFGS (\textbf{L-BFGS}) \cite{lbfgs}, \textbf{CMA-ES} \cite{cmaes}, and simulated annealing (\textbf{Anneal}) \cite{simanneal}. We also compare our method against the more recently introduced algorithms \textbf{TuRBO}-qEI \citep{turbo}, \textbf{COM} \citep{coms}, \textbf{RoMA} \citep{roma}, \textbf{BDI} \citep{bdi}, \textbf{DDOM} \citep{ddom}, \textbf{BONET} \citep{bonet}, \textbf{ExPT} \citep{expt}, \textbf{ROMO} \citep{romo}, and \textbf{BootGen} \citep{bootgen}. Because BootGen is proposed by \cite{bootgen} as an optimization method specifically for biological sequence design, we only assess this baseline method on the five relevant tasks in our evaluation suite.

\textbf{Conditional MBO Tasks.} To our knowledge, prior work in conditional model-based optimization is limited, and so previously reported algorithms are not equipped to solve such tasks out-of-the-box. \citet{romo} explore such tasks in their work, but primarily focus on conditional tasks that are built by arbitrarily fixing certain design dimensions from unconstrained problems, which are not representative of true conditional optimization problems in the real world. In our work, we introduce the Warfarin task to assess methods on their ability to design an optimal therapeutic drug regiment \textit{conditioned} on a fixed patient state and lab values. To assess existing methods on this task, we implement conditional proxies of all baselines employing a first-order optimization schema via \textit{partial} gradient ascent to only update the warfarin dose dimension while leaving the patient attribute conditional dimensions unchanged. Conditional BO-based methods are implemented by fitting separate Gaussian processes for each patient. In conditional DDOM, we exchange the algorithm's diffusion model-based backbone with a \textit{conditional} score-based diffusion model \citep{gu2023}.

Of note, the BONET algorithm \citep{bonet} requires multiple observations for any given patient to construct synthetic optimization trajectories. However, the key challenge in conditional MBO is that each condition (i.e., patient) has \textit{no} past observations (i.e., warfarin doses), and instead relies on learning from offline datasets constructed from different permutations of condition values 
As a result, the BONET algorithm is unable to be evaluated on conditional MBO tasks.

\subsection{Main Results}
\label{subsection:results}
Scoring of one-shot optimization candidates is shown in \textbf{Table \ref{table:results-top-1}}. Across all eight assessed tasks spanning a wide range of scientific domains, GABO with our aSCR algorithm achieved the best average rank of \textbf{3.8} when compared to other existing methods (next best is 5.5). Furthermore, GABO was able to propose top $k=1$ candidate designs that outperform the best design in the pre-existing offline dataset for 6 of the 8 tasks--greater than any of the other methods assessed. If a larger oracle evaluation budget is available (i.e., $k=128$), GABO with aSCR performs even better, achieving the best average rank of \textbf{3.0} (next best is 4.6). GABO is also the best algorithm on 3 of the 8 tasks and second best on 2 tasks according to this evaluation metric. Altogether, our results suggest that GABO is a promising method for proposing optimal design candidates in offline MBO.

Importantly, our aSCR algorithm improves upon both the na\"{i}ve BO-qEI and Grad. Ascent parent optimizers assessed. GABO outperforms both baseline BO-based optimization methods in our evaluation suite: BO (TuRBO) only achieves a rank of 8.8 (9.0) on the top $k=1$ evaluation metric and a rank of 6.6 (7.4) on the top $k=128$ metric. Similarly, GAGA scores an average rank of 7.4 (7.6) on the top $k=1$ ($k=128$) evaluation metric; by leveraging aSCR, GAGA outperforms its base parent optimizer (Grad. Ascent), which only achieves an average rank of 9.0 and 11.0 on the same two evaluation metrics, respectively. Our results show that using aSCR to adaptively penalize the objective of two popular optimization methods can improve their offline performance.

\begin{table*}[ht]
\caption{\textbf{Constrained Budget ($k=1$) Oracle Evaluation}$\quad$ Each method proposes a single design that is evaluated using the oracle function to report the final score (higher is better) across 10 random seeds reported as mean $\pm$ standard deviation. $\mathcal{D}$ (best) reports the top oracle value in the task dataset. Each of the MBO methods are ranked by their mean one-shot oracle score, and the average rank (lower is better) across all eight tasks is reported in the final table column. \textbf{Bold} (\underline{Underlined}) entries indicate the best (second best) entry in the column. $^*$Denotes the life sciences-related discrete MBO tasks from Design-Bench \citep{design-bench}.\vspace{-2ex}}
\label{table:results-top-1}
\begin{center}
\begin{small}
\resizebox{\textwidth}{!}{\begin{tabular}{rccccccccc}
\toprule
\textbf{Method} & \textbf{Branin} & \textbf{LogP} & \textbf{TF-Bind-8$^*$} & \textbf{GFP$^*$} & \textbf{UTR$^*$} & \textbf{ChEMBL$^*$} & \textbf{D'Kitty} & \textbf{Warfarin} & \textbf{Rank}\\
\midrule
$\mathcal{D}$ (best) & -13.0 & 11.3 & 0.439 & 3.53 & 7.12 & 0.61 & 0.88 & -0.19 $\pm$ 1.96 & --- \\
\midrule
Grad. & -245.1 $\pm$ 81.3 & -5.37 $\pm$ 1.44 & 0.429 $\pm$ 0.023 & 3.18 $\pm$ 0.88 & 6.82 $\pm$ 0.21 & -1.95 $\pm$ 0.00 & 0.57 $\pm$ 0.19 & \underline{0.86 $\pm$ 1.09} & 9.0\\
L-BFGS & -29.6 $\pm$ 0.0 & 3.82 $\pm$ 32.6 & 0.527 $\pm$ 0.140 & 3.51 $\pm$ 0.70 & 6.48 $\pm$ 1.20 & -1.95 $\pm$ 0.00 & 0.31 $\pm$ 0.00 & 0.73 $\pm$ 1.83 & 8.5\\
CMA-ES & -8.6 $\pm$ 3.6 & 5.04 $\pm$ 6.83 & 0.438 $\pm$ 0.131 & 1.43 $\pm$ 0.00 & 6.39 $\pm$ 0.11 & -1.95 $\pm$ 0.00 & 0.31 $\pm$ 0.00 & -25.0 $\pm$ 150 & 10.6\\
Anneal & -9.6 $\pm$ 1.5 & 8.76 $\pm$ 0.15 & \underline{0.807 $\pm$ 0.094} & \underline{3.64 $\pm$ 0.03} & 5.01 $\pm$ 0.31 & -1.95 $\pm$ 0.00 & 0.55 $\pm$ 0.18 & \textbf{0.91 $\pm$ 0.08} & 6.8\\
BO & -11.0 $\pm$ 7.8 & -52.5 $\pm$ 88.8 & 0.586 $\pm$ 0.193 & 1.43 $\pm$ 0.00 & 5.65 $\pm$ 1.30 & 0.59 $\pm$ 0.10 & 0.61 $\pm$ 0.15 & 0.16 $\pm$ 1.67 & 8.8\\
TuRBO & -21.0 $\pm$ 5.1 & -45.1 $\pm$ 93.8 & 0.564 $\pm$ 0.194 & 1.43 $\pm$ 0.00 & 6.53 $\pm$ 1.19 & \textbf{0.65 $\pm$ 0.00} & 0.44 $\pm$ 0.18 & 0.05 $\pm$ 0.11 & 9.0\\
BONET & -26.1 $\pm$ 0.9 & 10.8 $\pm$ 0.33 & 0.282 $\pm$ 0.000 & \textbf{3.74 $\pm$ 0.00} & \textbf{9.12 $\pm$ 0.07} & 0.55 $\pm$ 0.13 & 0.78 $\pm$ 0.00 & --- & 5.7\\
DDOM & -6677 $\pm$ 6360 & -4.23 $\pm$ 1.28 & 0.460 $\pm$ 0.030 & 1.43 $\pm$ 0.00 & 5.56 $\pm$ 0.02 & 0.54 $\pm$ 0.15 & 0.51 $\pm$ 0.20 & -0.32 $\pm$ 0.40 & 11.1\\
COM & -3099 $\pm$ 32.6 & \textbf{30.8 $\pm$ 19.5} & 0.439 $\pm$ 0.000 & 3.62 $\pm$ 0.00 & 6.65 $\pm$ 0.43 & \underline{0.63 $\pm$ 0.01} & \textbf{0.90 $\pm$ 0.02} & 0.72 $\pm$ 0.97 & \underline{5.5}\\
RoMA & -32.7 $\pm$ 18.4 & 6.37 $\pm$ 1.39 & 0.433 $\pm$ 0.040 & 3.37 $\pm$ 0.27 & 6.66 $\pm$ 0.98 & 0.50 $\pm$ 0.14 & 0.30 $\pm$ 0.27 & -0.70 $\pm$ 0.02 & 9.4\\
BDI & -1050 $\pm$ 0.0 & -0.20 $\pm$ 0.00 & 0.311 $\pm$ 0.000 & 3.26 $\pm$ 0.82 & 5.61 $\pm$ 0.00 & 0.48 $\pm$ 0.00 & 0.67 $\pm$ 0.00 & -24.8 $\pm$ 233 & 10.8\\
ExPT & -57.2 $\pm$ 38.6 & -15.9 $\pm$ 24.1 & 0.571 $\pm$ 0.076 & 1.43 $\pm$ 0.00 & 6.77 $\pm$ 1.38 & 0.56 $\pm$ 0.06 & 0.66 $\pm$ 0.20 & -34.6 $\pm$ 61.4 & 9.1\\
BootGen & --- & -13.0 $\pm$ 15.1 & \textbf{0.942 $\pm$ 0.022} & 3.10 $\pm$ 0.73 & \underline{8.30 $\pm$ 0.93} & 0.59 $\pm$ 0.07 & --- & --- & 6.2\\
ROMO & -2614 $\pm$ 739.9 & -20.5 $\pm$ 19.2 & 0.382 $\pm$ 0.203 & 3.55 $\pm$ 0.13 & 5.73 $\pm$ 1.42 & \textbf{0.65 $\pm$ 0.00} & 0.64 $\pm$ 0.27 & -0.71 $\pm$ 2.10 & 9.6\\
\midrule
\textbf{GAGA} & \underline{-2.9 $\pm$ 2.2} & -68.6 $\pm$ 109.8 & 0.571 $\pm$ 0.120 & \textbf{3.74 $\pm$ 0.00} & 5.89 $\pm$ 1.42 & -1.95 $\pm$ 0.00 & \underline{0.89 $\pm$ 0.00} & 0.01 $\pm$ 0.14 & 7.4\\
\textbf{GABO} & \textbf{-2.6 $\pm$ 1.1} & \underline{21.3 $\pm$ 33.2} & 0.570 $\pm$ 0.131 & 3.60 $\pm$ 0.40 & 7.51 $\pm$ 0.39 & 0.60 $\pm$ 0.07 & 0.71 $\pm$ 0.01 & 0.60 $\pm$ 1.80 & \textbf{3.8}\\
\bottomrule
\end{tabular}}
\end{small}
\end{center}
\vspace{-4ex}
\end{table*}

\begin{table*}[ht]
\caption{\textbf{Relaxed Budget ($k=128$) Oracle Evaluation}$\quad$ Each method now proposes 128 designs that are evaluated using the oracle function, and maximum score out of these 128 designs is reported below (averaged across 10 random seeds and reported as mean $\pm$ standard deviation). $\mathcal{D}$ (best) reports the top oracle value in the task dataset. Each of the MBO methods are ranked by their mean $k=128$-shot oracle score, and the average rank (lower is better) across all eight tasks is reported in the final table column. \textbf{Bold} (\underline{Underlined}) entries indicate the best (second best) entry in the column. $^*$Denotes the life sciences-related discrete MBO tasks from Design-Bench \citep{design-bench}.\vspace{-2ex}}
\label{table:results-top-128}
\begin{center}
\begin{small}
\resizebox{\textwidth}{!}{\begin{tabular}{rccccccccc}
\toprule
\textbf{Method} & \textbf{Branin} & \textbf{LogP} & \textbf{TF-Bind-8$^*$} & \textbf{GFP$^*$} & \textbf{UTR$^*$} & \textbf{ChEMBL$^*$} & \textbf{D'Kitty} & \textbf{Warfarin} & \textbf{Rank}\\
\midrule
$\mathcal{D}$ (best) & -13.0 & 11.3 & 0.439 & 3.53 & 7.12 & 0.61 & 0.88 & -0.19 $\pm$ 1.96 & --- \\
\midrule
Grad. & -115.3 $\pm$ 20.8 & -5.14 $\pm$ 1.70 & \underline{0.977 $\pm$ 0.025} & 3.49 $\pm$ 0.69 & 7.38 $\pm$ 0.15 & -1.95 $\pm$ 0.00 & 0.87 $\pm$ 0.02 & 0.86 $\pm$ 1.08 & 11.0\\
L-BFGS & -4.0 $\pm$ 0.0 & 42.8 $\pm$ 9.44 & 0.633 $\pm$ 0.140 & \textbf{3.74 $\pm$ 0.00} & 7.51 $\pm$ 0.39 & -1.95 $\pm$ 0.00 & 0.31 $\pm$ 0.00 & 0.75 $\pm$ 1.67 & 10.1\\
CMA-ES & -4.3 $\pm$ 1.7 & 47.6 $\pm$ 5.46 & 0.810 $\pm$ 0.235 & \textbf{3.74 $\pm$ 0.00} & 7.40 $\pm$ 0.32 & -1.95 $\pm$ 0.00 & 0.74 $\pm$ 0.00 & -8.62 $\pm$ 63.8 & 9.8\\
Anneal & -7.4 $\pm$ 2.8 & 11.3 $\pm$ 0.00 & 0.890 $\pm$ 0.035 & \underline{3.72 $\pm$ 0.00} & 7.96 $\pm$ 0.22 & -1.95 $\pm$ 0.00 & 0.88 $\pm$ 0.00 & 0.97 $\pm$ 0.08 & 9.3\\
BO & \textbf{-0.4 $\pm$ 0.0} & \textbf{135.3 $\pm$ 16.0} & 0.942 $\pm$ 0.025 & 2.26 $\pm$ 1.03 & 8.26 $\pm$ 0.09 & 0.67 $\pm$ 0.00 & 0.72 $\pm$ 0.00 & 0.93 $\pm$ 0.11 & 6.6\\
TuRBO & -0.7 $\pm$ 0.4 & 59.7 $\pm$ 51.3 & 0.895 $\pm$ 0.049 & 1.89 $\pm$ 0.92 & 8.26 $\pm$ 0.11 & 0.67 $\pm$ 0.01 & 0.72 $\pm$ 0.00 & \underline{0.99 $\pm$ 0.01} & 7.4\\
BONET & -26.0 $\pm$ 0.9 & 11.7 $\pm$ 0.38 & 0.951 $\pm$ 0.035 & \textbf{3.74 $\pm$ 0.00} & \underline{9.13 $\pm$ 0.08} & 0.67 $\pm$ 0.01 & 0.95 $\pm$ 0.01 & --- & 5.6\\
DDOM & -18.4 $\pm$ 29.8 & -2.16 $\pm$ 0.60 & 0.936 $\pm$ 0.051 & 1.44 $\pm$ 0.00 & 8.30 $\pm$ 0.33 & 0.66 $\pm$ 0.01 & 0.89 $\pm$ 0.01 & \textbf{1.00 $\pm$ 0.00} & 8.4\\
COM & -1981 $\pm$ 224.5 & 42.0 $\pm$ 16.9 & 0.902 $\pm$ 0.056 & 3.62 $\pm$ 0.00 & 8.18 $\pm$ 0.00 & 0.64 $\pm$ 0.01 & 0.95 $\pm$ 0.02 & 0.77 $\pm$ 0.86 & 8.5\\
RoMA & -4.8 $\pm$ 3.0 & 10.8 $\pm$ 0.78 & 0.760 $\pm$ 0.113 & \textbf{3.74 $\pm$ 0.00} & 8.12 $\pm$ 0.09 & \underline{0.69 $\pm$ 0.03} & \textbf{1.02 $\pm$ 0.04} & 0.67 $\pm$ 0.05 & 7.8\\
BDI & -65.0 $\pm$ 51.3 & 1.52 $\pm$ 5.79 & 0.735 $\pm$ 0.086 & 3.61 $\pm$ 0.05 & 6.31 $\pm$ 0.00 & 0.50 $\pm$ 0.12 & 0.94 $\pm$ 0.01 & -5.07 $\pm$ 21.0 & 11.8\\
ExPT & -1.7 $\pm$ 1.0 & -6.48 $\pm$ 4.58 & 0.927 $\pm$ 0.095 & \textbf{3.74 $\pm$ 0.00} & 8.13 $\pm$ 0.09 & 0.68 $\pm$ 0.04 & \underline{0.97 $\pm$ 0.01} & 0.96 $\pm$ 0.05 & 6.5\\
BootGen & --- & 8.10 $\pm$ 3.31 & \textbf{0.979 $\pm$ 0.002} & \textbf{3.74 $\pm$ 0.00} & \textbf{10.5 $\pm$ 0.95} & 0.68 $\pm$ 0.00 & --- & --- & \underline{4.6}\\
ROMO & -2367 $\pm$ 787.5 & -6.05 $\pm$ 14.5 & 0.572 $\pm$ 0.202 & 3.67 $\pm$ 0.03 & 6.94 $\pm$ 1.07 & 0.65 $\pm$ 0.00 & 0.90 $\pm$ 0.02 & 0.76 $\pm$ 1.91 & 12.1\\
\midrule
\textbf{GAGA} & -1.0 $\pm$ 0.2 & 14.1 $\pm$ 25.0 & 0.722 $\pm$ 0.091 & \textbf{3.74 $\pm$ 0.00} & 7.98 $\pm$ 0.36 & -1.95 $\pm$ 0.00 & 0.90 $\pm$ 0.01 & 0.95 $\pm$ 0.07 & 7.6\\
\textbf{GABO} & \underline{-0.5 $\pm$ 0.1} & \underline{122.1 $\pm$ 20.6} & 0.954 $\pm$ 0.025 & \textbf{3.74 $\pm$ 0.00} & 8.36 $\pm$ 0.08 & \textbf{0.70 $\pm$ 0.01} & 0.72 $\pm$ 0.00 & \textbf{1.00 $\pm$ 0.03} & \textbf{3.0} \\
\bottomrule
\end{tabular}}
\end{small}
\end{center}
\vspace{-4ex}
\end{table*}

\begin{figure*}[ht]
    \centering
    \includegraphics[width=\textwidth]{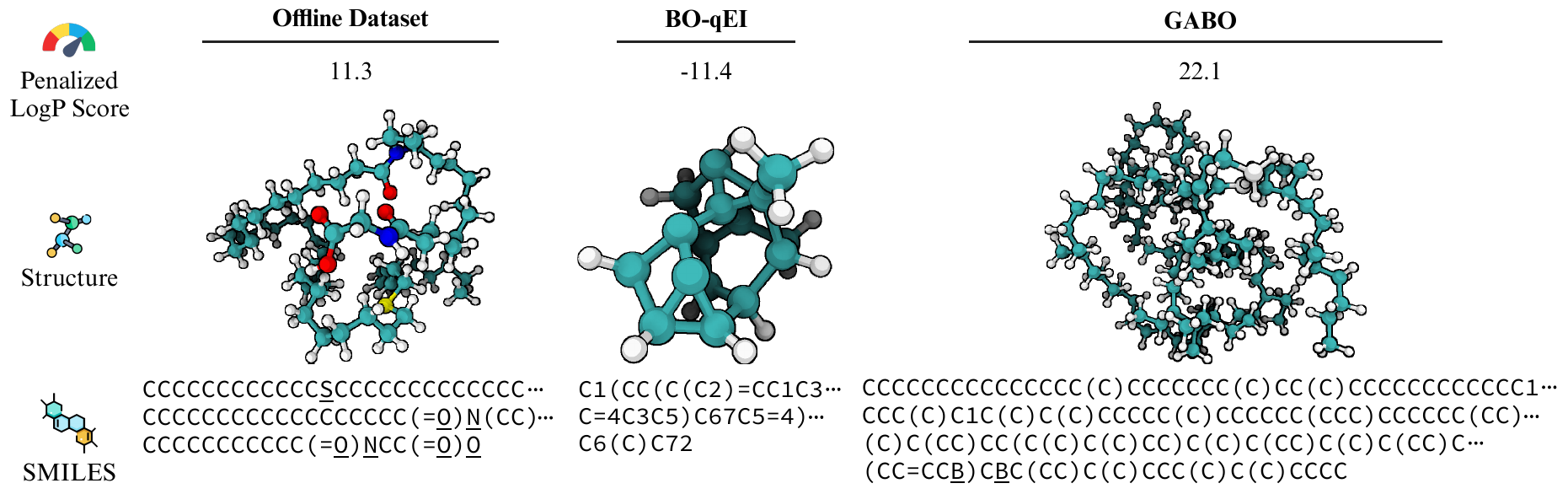}
    \vskip -0.1in
    \caption{\textbf{Penalized LogP Score Maximization Sample Candidate Designs}$\quad$ (\textbf{Left}) The molecule with the highest penalized LogP score of 11.3 in the offline dataset. Separately, we show the 100th percentile candidate molecules according to the surrogate objective generated from (\textbf{Middle}) vanilla BO-qEI and (\textbf{Right}) GABO. Teal- (white-) colored atoms are carbon (hydrogen). Non-hydrocarbon atoms are underlined in the SMILES \citep{smiles} string representations of the molecules.}
    \label{fig:examples}
    \vskip -0.05in
\end{figure*}

\textbf{Qualitative Evaluation: Penalized LogP Task.} We evaluate GABO against na\"{i}ve BO-qEI for the \textbf{LogP} task by inspecting the three-dimensional chemical structures of the top-scoring candidate molecules. As a general principle, molecules that are associated with high Penalized LogP scores are hydrophobic with minimal ring structures and therefore often feature long hydrocarbon backbones \citep{logp}. In \textbf{Figure \ref{fig:examples}}, we see that BO-qEI using the unconstrained surrogate objective generates a candidate molecule of hydrogen and carbon atoms. However, the proposed candidate includes two rings in its structure, resulting in a suboptimal oracle Penalized LogP score.

We hypothesize that this may be due to a lack of ring-containing example molecules in the offline dataset, as only 6.7\% (2.7\%) of observed molecules contain at least one (two) carbon ring(s). As a result, the surrogate objective model estimator returns more inaccurate Penalized LogP estimates for input ring-containing structures (surrogate model root mean squared error (RMSE) = 25.5 for offline dataset molecules with at least 2 rings; RMSE = 16.5 for those with at least 1 ring; and RMSE = 4.6 for those with at least 0 rings), leading to sub-par BO-qEI optimization performance as the unconstrained algorithm extrapolates against the surrogate to find ``optimal'' molecules that are out-of-distribution. In contrast, GABO generates a candidate molecule with a long hydrocarbon backbone and \textit{no} rings, resulting in a penalized logP score of 22.1\textemdash greater than the best observed value in the offline dataset for the task.

\textbf{Ablation Experiments.} Taking inspiration from \citep{coms}, it is possible to utilize our SCR algorithm in GABO \textit{without} dynamically computing $\alpha$ (and hence the Lagrange multiplier $\lambda$). To better characterize the utility of aSCR, we ablate \textbf{Algorithm \ref{alg:ascr}} by treating $\lambda$ instead as a hand-tunable constant hyperparameter, and test our method using different values of $\lambda=\alpha/(1-\alpha)$ (\textbf{Table \ref{table:ablation}}). Setting $\alpha=0$ (i.e., $\lambda=0$) corresponds to na\"{i}ve BO against the unconstrained surrogate model, while $\alpha=1$ (i.e., $\lambda\rightarrow \infty$) is equivalent to a WGAN-like policy. Evaluating constant values of $\alpha$ ranging from 0 to 1, we find that there is no consistently optimal constant value for all eight optimization tasks. In contrast, our method achieves an average rank of \textbf{1.9} (\textbf{2.4}) on the top-1 (top-128) evaluation metric, and is one of the top two methods when compared to the ablations for at least five of the eight tasks. These results suggest that the `adaptive' nature of aSCR is an important component in solving the constrained optimization problem in (\ref{eq:constrained-opt}).

\begin{table*}[ht]
\vspace{-1.5ex}
\caption{\textbf{GABO Adaptive SCR Ablation Study}$\quad$ One-shot ($k=1$) and few-shot ($k=128$) oracle evaluations averaged across 10 random seeds reported as mean $\pm$ standard deviation. $\mathcal{D}$ (best) reports the top oracle value in the task dataset.\vspace{-2ex}}
\label{table:ablation}
\begin{center}
\begin{small}
\resizebox{\textwidth}{!}{\begin{tabular}{rcccccccccc}
\toprule
\textbf{Top-1} & \textbf{Branin} & \textbf{LogP} & \textbf{TF-Bind-8$^*$} & \textbf{GFP$^*$} & \textbf{UTR$^*$} & \textbf{ChEMBL$^*$} & \textbf{D'Kitty} & \textbf{Warfarin} & \textbf{Rank}\\
\midrule
$\mathcal{D}$ (best) & -13.0 & 11.3 & 0.439 & 3.53 & 7.12 & 0.61 & 0.88 & -0.19 $\pm$ 1.96 & --- \\
\midrule
$\alpha=0.0$ & -11.0 $\pm$ 7.8 & -52.5 $\pm$ 88.8 & 0.586 $\pm$ 0.193 & 1.43 $\pm$ 0.00 & 5.65 $\pm$ 1.30 & 0.59 $\pm$ 0.10 & 0.61 $\pm$ 0.15 & \underline{0.16 $\pm$ 1.67} & 4.5 \\
$\alpha=0.2$ & -9.8 $\pm$ 3.9 & \underline{-4.39 $\pm$ 60.7} & 0.535 $\pm$ 0.110 & 1.43 $\pm$ 0.00 & 4.69 $\pm$ 1.44 & \underline{0.63 $\pm$ 0.03} & 0.61 $\pm$ 0.15 & \underline{0.16 $\pm$ 1.79} & 3.9 \\
$\alpha=0.5$ & -7.9 $\pm$ 6.6 & -83.9 $\pm$ 166.3 & \underline{0.601 $\pm$ 0.212} & 1.43 $\pm$ 0.00 & 5.69 $\pm$ 1.51 & \underline{0.63 $\pm$ 0.04} & \underline{0.66 $\pm$ 0.12} & \underline{0.16 $\pm$ 1.79} & 3.6 \\
$\alpha=0.8$ & \underline{-5.2 $\pm$ 3.1} & -43.3 $\pm$ 170.0 & \textbf{0.654 $\pm$ 0.218} & 1.66 $\pm$ 0.69 & \underline{6.49 $\pm$ 1.20} & \textbf{0.64 $\pm$ 0.02} & \textbf{0.71 $\pm$ 0.01} & \underline{0.16 $\pm$ 1.80} & \underline{2.4} \\
$\alpha=1.0$ & -99.5 $\pm$ 61.2 & -46.8 $\pm$ 114.3 & 0.454 $\pm$ 0.120 & \textbf{3.74 $\pm$ 0.01} & 5.26 $\pm$ 2.35 & 0.52 $\pm$ 0.16 & 0.62 $\pm$ 0.15 & -9.04 $\pm$ 57.3 & 4.8 \\
\midrule
\textbf{aSCR} & \textbf{-2.6 $\pm$ 1.1} & \textbf{21.3 $\pm$ 33.2} & 0.570 $\pm$ 0.131 & \underline{3.60 $\pm$ 0.40} & \textbf{7.51 $\pm$ 0.39} & 0.60 $\pm$ 0.07 & \textbf{0.71 $\pm$ 0.01} & \textbf{0.60 $\pm$ 1.80} & \textbf{1.9} \\
\bottomrule
 & \\
\toprule
\textbf{Top-128} & \textbf{Branin} & \textbf{LogP} & \textbf{TF-Bind-8$^*$} & \textbf{GFP$^*$} & \textbf{UTR$^*$} & \textbf{ChEMBL$^*$} & \textbf{D'Kitty} & \textbf{Warfarin} & \textbf{Rank}\\
\midrule
$\mathcal{D}$ (best) & -13.0 & 11.3 & 0.439 & 3.53 & 7.12 & 0.61 & 0.88 & -0.19 $\pm$ 1.96 & --- \\
\midrule
$\alpha=0.0$ & \textbf{-0.4 $\pm$ 0.0} & \underline{135.3 $\pm$ 16.0} & 0.942 $\pm$ 0.025 & 2.26 $\pm$ 1.03 & 8.26 $\pm$ 0.09 & 0.67 $\pm$ 0.00 & 0.72 $\pm$ 0.00 & 0.93 $\pm$ 0.11 & 4.3 \\
$\alpha=0.2$ & \textbf{-0.4 $\pm$ 0.1} & 121.8 $\pm$ 20.6 & 0.925 $\pm$ 0.029 & 3.01 $\pm$ 1.04 & 8.20 $\pm$ 0.10 & 0.67 $\pm$ 0.01 & 0.72 $\pm$ 0.00 & \textbf{1.00 $\pm$ 0.00} & 4.8 \\
$\alpha=0.5$ & \textbf{-0.4 $\pm$ 0.0} & 127.7 $\pm$ 23.1 & \underline{0.944 $\pm$ 0.040} & \underline{3.49 $\pm$ 0.69} & 8.29 $\pm$ 0.08 & 0.67 $\pm$ 0.01 & 0.72 $\pm$ 0.00 & \textbf{1.00 $\pm$ 0.00} & \underline{2.9} \\
$\alpha=0.8$ & \textbf{-0.4 $\pm$ 0.0} & 104.5 $\pm$ 31.8 & 0.933 $\pm$ 0.036 & \textbf{3.74 $\pm$ 0.00} & \underline{8.38 $\pm$ 0.11} & 0.67 $\pm$ 0.02 & 0.72 $\pm$ 0.00 & \textbf{1.00 $\pm$ 0.00} & 3.4 \\
$\alpha=1.0$ & -2.2 $\pm$ 1.4 & \textbf{142.3 $\pm$ 2.41} & 0.906 $\pm$ 0.061 & \textbf{3.74 $\pm$ 0.00} & \textbf{8.54 $\pm$ 0.08} & \underline{0.68 $\pm$ 0.01} & 0.72 $\pm$ 0.00 & \underline{0.99 $\pm$ 0.04} & 3.4 \\
\midrule
\textbf{aSCR} & \underline{-0.5 $\pm$ 0.1} & 122.1 $\pm$ 20.6 & \textbf{0.954 $\pm$ 0.025} & \textbf{3.74 $\pm$ 0.00} & 8.36 $\pm$ 0.08 & \textbf{0.70 $\pm$ 0.01} & 0.72 $\pm$ 0.00 & \textbf{1.00 $\pm$ 0.03} & \textbf{2.4} \\
\bottomrule
\end{tabular}}
\end{small}
\end{center}
\vspace{-4ex}
\end{table*}

Of note, the top designs found across different constant values of $\alpha$ can be very similar for certain tasks. This reflects the inherent challenge in developing task-agnostic methods for policy regularization\textemdash if the magnitudes of the unconstrained objective and regularization function vastly differ, then constant values of $\alpha$ may over- or under- constrain the objective. Adaptive SCR overcomes this problem by dynamically setting $\alpha$ as an implicit function of prior observations.

\section{Conclusion}
\label{section:conclusion}

We propose \textbf{adaptive source critic regularization (aSCR)} to solve the problem of off-distribution objective evaluation in offline MBO. When leveraged with vanilla Bayesian optimization, aSCR outperforms baseline methods to achieve an average rank of \textbf{3.8} (\textbf{3.0}) in one-shot $k=1$ (few-shot $k=128$) oracle evaluation, and most consistently proposes designs better than the offline dataset.

\textbf{Limitations.}
One limitation of aSCR is that our algorithm requires preexisting knowledge of the prior distribution over the input space in order to be computationally tractable. While we have focused our experimental evaluation on tasks amenable to imposed latent space priors, further work is needed to adapt aSCR to any arbitrary configuration space. Future work may also extend aSCR to improve parent optimization methods more sophisticated than BO-qEI and Gradient Ascent explored herein.

\textbf{Impact Statement.} Offline policy optimization methods, such as those discussed in this work, have the potential to benefit society. Such examples may include helping develop more effective drugs and individualizing patient therapies. However, as with any real-world algorithm, these methods can also be leveraged to generate potentially harmful design candidates. Careful oversight by domain experts and researchers is required to ensure that the contributions proposed herein are used for social good.

\section*{Funding Disclosure and Acknowledgements}
The authors thank Pratik Chaudhari at the University of Pennsylvania and the anonymous NeurIPS peer reviewers for their thoughtful comments, feedback, and discussion regarding this work. MSY is supported by NIH F30 MD020264. YZ and JRG are supported by NSF award IIS-2145644. JCG is supported by NIH R01 EB031722. OB is supported by NSF Award CCF-1917852.

\bibliography{gabo}
\bibliographystyle{icml2024}

\newpage
\section*{NeurIPS Paper Checklist}

\begin{enumerate}

\item {\bf Claims}
    \item[] Question: Do the main claims made in the abstract and introduction accurately reflect the paper's contributions and scope?
    \item[] Answer: \answerYes{} 
    \item[] Justification: The main claims made in the abstract and introduction are supported by the experimental results presented. The introduction includes the contributions made in the paper, and important assumptions and limitations are included where relevant.
    \item[] Guidelines:
    \begin{itemize}
        \item The answer NA means that the abstract and introduction do not include the claims made in the paper.
        \item The abstract and/or introduction should clearly state the claims made, including the contributions made in the paper and important assumptions and limitations. A No or NA answer to this question will not be perceived well by the reviewers. 
        \item The claims made should match theoretical and experimental results, and reflect how much the results can be expected to generalize to other settings. 
        \item It is fine to include aspirational goals as motivation as long as it is clear that these goals are not attained by the paper. 
    \end{itemize}

\item {\bf Limitations}
    \item[] Question: Does the paper discuss the limitations of the work performed by the authors?
    \item[] Answer: \answerYes{}
    \item[] Justification: Please see Section \ref{section:conclusion} in the main text for a focused `Limitations' section and relevant discussion.
    \item[] Guidelines:
    \begin{itemize}
        \item The answer NA means that the paper has no limitation while the answer No means that the paper has limitations, but those are not discussed in the paper. 
        \item The authors are encouraged to create a separate "Limitations" section in their paper.
        \item The paper should point out any strong assumptions and how robust the results are to violations of these assumptions (e.g., independence assumptions, noiseless settings, model well-specification, asymptotic approximations only holding locally). The authors should reflect on how these assumptions might be violated in practice and what the implications would be.
        \item The authors should reflect on the scope of the claims made, e.g., if the approach was only tested on a few datasets or with a few runs. In general, empirical results often depend on implicit assumptions, which should be articulated.
        \item The authors should reflect on the factors that influence the performance of the approach. For example, a facial recognition algorithm may perform poorly when image resolution is low or images are taken in low lighting. Or a speech-to-text system might not be used reliably to provide closed captions for online lectures because it fails to handle technical jargon.
        \item The authors should discuss the computational efficiency of the proposed algorithms and how they scale with dataset size.
        \item If applicable, the authors should discuss possible limitations of their approach to address problems of privacy and fairness.
        \item While the authors might fear that complete honesty about limitations might be used by reviewers as grounds for rejection, a worse outcome might be that reviewers discover limitations that aren't acknowledged in the paper. The authors should use their best judgment and recognize that individual actions in favor of transparency play an important role in developing norms that preserve the integrity of the community. Reviewers will be specifically instructed to not penalize honesty concerning limitations.
    \end{itemize}

\item {\bf Theory Assumptions and Proofs}
    \item[] Question: For each theoretical result, does the paper provide the full set of assumptions and a complete (and correct) proof?
    \item[] Answer: \answerNA{} 
    \item[] Justification: The paper does not include theoretical results.
    \item[] Guidelines:
    \begin{itemize}
        \item The answer NA means that the paper does not include theoretical results. 
        \item All the theorems, formulas, and proofs in the paper should be numbered and cross-referenced.
        \item All assumptions should be clearly stated or referenced in the statement of any theorems.
        \item The proofs can either appear in the main paper or the supplemental material, but if they appear in the supplemental material, the authors are encouraged to provide a short proof sketch to provide intuition. 
        \item Inversely, any informal proof provided in the core of the paper should be complemented by formal proofs provided in appendix or supplemental material.
        \item Theorems and Lemmas that the proof relies upon should be properly referenced. 
    \end{itemize}

\item {\bf Experimental Result Reproducibility}
    \item[] Question: Does the paper fully disclose all the information needed to reproduce the main experimental results of the paper to the extent that it affects the main claims and/or conclusions of the paper (regardless of whether the code and data are provided or not)?
    \item[] Answer: \answerYes{} 
    \item[] Justification: The paper fully discloses all the relevant information needed to reproduce all reported experimental results of the paper. We have made our code required to reproduce our experimental results available in the Supplementary Material ZIP file.
    \item[] Guidelines:
    \begin{itemize}
        \item The answer NA means that the paper does not include experiments.
        \item If the paper includes experiments, a No answer to this question will not be perceived well by the reviewers: Making the paper reproducible is important, regardless of whether the code and data are provided or not.
        \item If the contribution is a dataset and/or model, the authors should describe the steps taken to make their results reproducible or verifiable. 
        \item Depending on the contribution, reproducibility can be accomplished in various ways. For example, if the contribution is a novel architecture, describing the architecture fully might suffice, or if the contribution is a specific model and empirical evaluation, it may be necessary to either make it possible for others to replicate the model with the same dataset, or provide access to the model. In general. releasing code and data is often one good way to accomplish this, but reproducibility can also be provided via detailed instructions for how to replicate the results, access to a hosted model (e.g., in the case of a large language model), releasing of a model checkpoint, or other means that are appropriate to the research performed.
        \item While NeurIPS does not require releasing code, the conference does require all submissions to provide some reasonable avenue for reproducibility, which may depend on the nature of the contribution. For example
        \begin{enumerate}
            \item If the contribution is primarily a new algorithm, the paper should make it clear how to reproduce that algorithm.
            \item If the contribution is primarily a new model architecture, the paper should describe the architecture clearly and fully.
            \item If the contribution is a new model (e.g., a large language model), then there should either be a way to access this model for reproducing the results or a way to reproduce the model (e.g., with an open-source dataset or instructions for how to construct the dataset).
            \item We recognize that reproducibility may be tricky in some cases, in which case authors are welcome to describe the particular way they provide for reproducibility. In the case of closed-source models, it may be that access to the model is limited in some way (e.g., to registered users), but it should be possible for other researchers to have some path to reproducing or verifying the results.
        \end{enumerate}
    \end{itemize}

\item {\bf Open access to data and code}
    \item[] Question: Does the paper provide open access to the data and code, with sufficient instructions to faithfully reproduce the main experimental results, as described in supplemental material?
    \item[] Answer: \answerYes{}
    \item[] Justification: All data and code associated with this paper is open access and includes sufficient instructions to faithfully reproduce the experimental results reported herein. We have made the data and code available in the Supplementary Material ZIP file. The public link to the GitHub repository containing the code will be included in the Abstract in the final version (currently not linked so as to comply with double-blind review). All datasets used herein are publicly available without any limitations in public accessibility. Our data and code adhere to the \href{https://nips.cc/public/guides/CodeSubmissionPolicy}{NeurIPS code and data submission guidelines}.
    \item[] Guidelines:
    \begin{itemize}
        \item The answer NA means that paper does not include experiments requiring code.
        \item Please see the NeurIPS code and data submission guidelines (\url{https://nips.cc/public/guides/CodeSubmissionPolicy}) for more details.
        \item While we encourage the release of code and data, we understand that this might not be possible, so “No” is an acceptable answer. Papers cannot be rejected simply for not including code, unless this is central to the contribution (e.g., for a new open-source benchmark).
        \item The instructions should contain the exact command and environment needed to run to reproduce the results. See the NeurIPS code and data submission guidelines (\url{https://nips.cc/public/guides/CodeSubmissionPolicy}) for more details.
        \item The authors should provide instructions on data access and preparation, including how to access the raw data, preprocessed data, intermediate data, and generated data, etc.
        \item The authors should provide scripts to reproduce all experimental results for the new proposed method and baselines. If only a subset of experiments are reproducible, they should state which ones are omitted from the script and why.
        \item At submission time, to preserve anonymity, the authors should release anonymized versions (if applicable).
        \item Providing as much information as possible in supplemental material (appended to the paper) is recommended, but including URLs to data and code is permitted.
    \end{itemize}

\item {\bf Experimental Setting/Details}
    \item[] Question: Does the paper specify all the training and test details (e.g., data splits, hyperparameters, how they were chosen, type of optimizer, etc.) necessary to understand the results?
    \item[] Answer: \answerYes{} 
    \item[] Justification: The experimental setting is clearly described in \textbf{Section \ref{subsection:eval}} and \textbf{Appendix \ref{appendix:a}}, and is also reproducible via the code released alongside the paper in the Supplementary Material ZIP file.
    \item[] Guidelines:
    \begin{itemize}
        \item The answer NA means that the paper does not include experiments.
        \item The experimental setting should be presented in the core of the paper to a level of detail that is necessary to appreciate the results and make sense of them.
        \item The full details can be provided either with the code, in appendix, or as supplemental material.
    \end{itemize}

\item {\bf Experiment Statistical Significance}
    \item[] Question: Does the paper report error bars suitably and correctly defined or other appropriate information about the statistical significance of the experiments?
    \item[] Answer: \answerYes{} 
    \item[] Justification: All statistics and results included in the paper are accompanied by confidence intervals. We clearly describe the factors of variability in the confidence intervals in \textbf{Section \ref{subsection:eval}} and \textbf{Appendix \ref{appendix:a}}. All additional details are included in relevant table captions.
    \item[] Guidelines:
    \begin{itemize}
        \item The answer NA means that the paper does not include experiments.
        \item The authors should answer "Yes" if the results are accompanied by error bars, confidence intervals, or statistical significance tests, at least for the experiments that support the main claims of the paper.
        \item The factors of variability that the error bars are capturing should be clearly stated (for example, train/test split, initialization, random drawing of some parameter, or overall run with given experimental conditions).
        \item The method for calculating the error bars should be explained (closed form formula, call to a library function, bootstrap, etc.)
        \item The assumptions made should be given (e.g., Normally distributed errors).
        \item It should be clear whether the error bar is the standard deviation or the standard error of the mean.
        \item It is OK to report 1-sigma error bars, but one should state it. The authors should preferably report a 2-sigma error bar than state that they have a 96\% CI, if the hypothesis of Normality of errors is not verified.
        \item For asymmetric distributions, the authors should be careful not to show in tables or figures symmetric error bars that would yield results that are out of range (e.g. negative error rates).
        \item If error bars are reported in tables or plots, The authors should explain in the text how they were calculated and reference the corresponding figures or tables in the text.
    \end{itemize}

\item {\bf Experiments Compute Resources}
    \item[] Question: For each experiment, does the paper provide sufficient information on the computer resources (type of compute workers, memory, time of execution) needed to reproduce the experiments?
    \item[] Answer: \answerYes{} 
    \item[] Justification: Information for the resources required to reproduce the experiments are included in \textbf{Section \ref{subsection:eval}} in the main paper. More specifically, all experiments were performed on a single internal cluster with 8 NVIDIA RTX A6000 GPUs. Any particular experimental configuration required no more than 24 hours to complete using our setup. The full research project did not require more compute than the experiments reported in the paper.
    \item[] Guidelines:
    \begin{itemize}
        \item The answer NA means that the paper does not include experiments.
        \item The paper should indicate the type of compute workers CPU or GPU, internal cluster, or cloud provider, including relevant memory and storage.
        \item The paper should provide the amount of compute required for each of the individual experimental runs as well as estimate the total compute. 
        \item The paper should disclose whether the full research project required more compute than the experiments reported in the paper (e.g., preliminary or failed experiments that didn't make it into the paper). 
    \end{itemize}
    
\item {\bf Code Of Ethics}
    \item[] Question: Does the research conducted in the paper conform, in every respect, with the NeurIPS Code of Ethics \url{https://neurips.cc/public/EthicsGuidelines}?
    \item[] Answer: \answerYes{}
    \item[] Justification: The authors have reviewed the NeurIPS Code of Ethics and assert that the research described herein conforms with the Code of Ethics in every respect.
    \item[] Guidelines:
    \begin{itemize}
        \item The answer NA means that the authors have not reviewed the NeurIPS Code of Ethics.
        \item If the authors answer No, they should explain the special circumstances that require a deviation from the Code of Ethics.
        \item The authors should make sure to preserve anonymity (e.g., if there is a special consideration due to laws or regulations in their jurisdiction).
    \end{itemize}

\item {\bf Broader Impacts}
    \item[] Question: Does the paper discuss both potential positive societal impacts and negative societal impacts of the work performed?
    \item[] Answer: \answerYes{}
    \item[] Justification: Please see the conclusion (i.e. \textbf{Section \ref{section:conclusion}}) in the main text for a focused `Impact Statement' subsection and relevant discussion on the potential societal impacts of our work.
    \item[] Guidelines:
    \begin{itemize}
        \item The answer NA means that there is no societal impact of the work performed.
        \item If the authors answer NA or No, they should explain why their work has no societal impact or why the paper does not address societal impact.
        \item Examples of negative societal impacts include potential malicious or unintended uses (e.g., disinformation, generating fake profiles, surveillance), fairness considerations (e.g., deployment of technologies that could make decisions that unfairly impact specific groups), privacy considerations, and security considerations.
        \item The conference expects that many papers will be foundational research and not tied to particular applications, let alone deployments. However, if there is a direct path to any negative applications, the authors should point it out. For example, it is legitimate to point out that an improvement in the quality of generative models could be used to generate deepfakes for disinformation. On the other hand, it is not needed to point out that a generic algorithm for optimizing neural networks could enable people to train models that generate Deepfakes faster.
        \item The authors should consider possible harms that could arise when the technology is being used as intended and functioning correctly, harms that could arise when the technology is being used as intended but gives incorrect results, and harms following from (intentional or unintentional) misuse of the technology.
        \item If there are negative societal impacts, the authors could also discuss possible mitigation strategies (e.g., gated release of models, providing defenses in addition to attacks, mechanisms for monitoring misuse, mechanisms to monitor how a system learns from feedback over time, improving the efficiency and accessibility of ML).
    \end{itemize}
    
\item {\bf Safeguards}
    \item[] Question: Does the paper describe safeguards that have been put in place for responsible release of data or models that have a high risk for misuse (e.g., pretrained language models, image generators, or scraped datasets)?
    \item[] Answer: \answerNA{}
    \item[] Justification: Our paper does not introduce any assets that have a high risk for misuse.
    \item[] Guidelines:
    \begin{itemize}
        \item The answer NA means that the paper poses no such risks.
        \item Released models that have a high risk for misuse or dual-use should be released with necessary safeguards to allow for controlled use of the model, for example by requiring that users adhere to usage guidelines or restrictions to access the model or implementing safety filters. 
        \item Datasets that have been scraped from the Internet could pose safety risks. The authors should describe how they avoided releasing unsafe images.
        \item We recognize that providing effective safeguards is challenging, and many papers do not require this, but we encourage authors to take this into account and make a best faith effort.
    \end{itemize}

\item {\bf Licenses for existing assets}
    \item[] Question: Are the creators or original owners of assets (e.g., code, data, models), used in the paper, properly credited and are the license and terms of use explicitly mentioned and properly respected?
    \item[] Answer: \answerYes{} 
    \item[] Justification: The only existing assets used in the paper are the seven MBO datasets used for experimental evaluation as described in \textbf{Section \ref{section:experimental}}. All of the relevant datasets are publicly available, and the references for each of the datasets are cited in the aforementioned section. The licenses associated with each of the seven datasets made available in each of the relevant citations are properly respected. There are no restrictions with respect to accessing any of the datasets used in the paper. No scraped data was used in this paper.
    \item[] Guidelines:
    \begin{itemize}
        \item The answer NA means that the paper does not use existing assets.
        \item The authors should cite the original paper that produced the code package or dataset.
        \item The authors should state which version of the asset is used and, if possible, include a URL.
        \item The name of the license (e.g., CC-BY 4.0) should be included for each asset.
        \item For scraped data from a particular source (e.g., website), the copyright and terms of service of that source should be provided.
        \item If assets are released, the license, copyright information, and terms of use in the package should be provided. For popular datasets, \url{paperswithcode.com/datasets} has curated licenses for some datasets. Their licensing guide can help determine the license of a dataset.
        \item For existing datasets that are re-packaged, both the original license and the license of the derived asset (if it has changed) should be provided.
        \item If this information is not available online, the authors are encouraged to reach out to the asset's creators.
    \end{itemize}

\item {\bf New Assets}
    \item[] Question: Are new assets introduced in the paper well documented and is the documentation provided alongside the assets?
    \item[] Answer: \answerYes{} 
    \item[] Justification: The only asset introduced in and released alongside the paper is the experimental code to reproduce the reported results. The repository containing the code is included in the Supplementary Material ZIP file. All datasets used for the experiments discussed herein are publicly available and available online via the appropriate references in this paper and via online links included in the Supplementary Material ZIP file.
    \item[] Guidelines:
    \begin{itemize}
        \item The answer NA means that the paper does not release new assets.
        \item Researchers should communicate the details of the dataset/code/model as part of their submissions via structured templates. This includes details about training, license, limitations, etc. 
        \item The paper should discuss whether and how consent was obtained from people whose asset is used.
        \item At submission time, remember to anonymize your assets (if applicable). You can either create an anonymized URL or include an anonymized zip file.
    \end{itemize}

\item {\bf Crowdsourcing and Research with Human Subjects}
    \item[] Question: For crowdsourcing experiments and research with human subjects, does the paper include the full text of instructions given to participants and screenshots, if applicable, as well as details about compensation (if any)? 
    \item[] Answer: \answerNA{}
    \item[] Justification: The research described herein does not involve crowdsourcing or research with human subjects.
    \item[] Guidelines:
    \begin{itemize}
        \item The answer NA means that the paper does not involve crowdsourcing nor research with human subjects.
        \item Including this information in the supplemental material is fine, but if the main contribution of the paper involves human subjects, then as much detail as possible should be included in the main paper. 
        \item According to the NeurIPS Code of Ethics, workers involved in data collection, curation, or other labor should be paid at least the minimum wage in the country of the data collector. 
    \end{itemize}

\item {\bf Institutional Review Board (IRB) Approvals or Equivalent for Research with Human Subjects}
    \item[] Question: Does the paper describe potential risks incurred by study participants, whether such risks were disclosed to the subjects, and whether Institutional Review Board (IRB) approvals (or an equivalent approval/review based on the requirements of your country or institution) were obtained?
    \item[] Answer: \answerNA{}
    \item[] Justification: The research described herein does not involve crowdsourcing or research with human subjects.
    \item[] Guidelines:
    \begin{itemize}
        \item The answer NA means that the paper does not involve crowdsourcing nor research with human subjects.
        \item Depending on the country in which research is conducted, IRB approval (or equivalent) may be required for any human subjects research. If you obtained IRB approval, you should clearly state this in the paper. 
        \item We recognize that the procedures for this may vary significantly between institutions and locations, and we expect authors to adhere to the NeurIPS Code of Ethics and the guidelines for their institution. 
        \item For initial submissions, do not include any information that would break anonymity (if applicable), such as the institution conducting the review.
    \end{itemize}

\end{enumerate}

\newpage
\appendix
\setcounter{table}{0}
\renewcommand{\thetable}{A\arabic{table}}
\onecolumn
\section{Additional Implementation Details}
\label{appendix:a}

\textbf{Oracle Functions.}
All oracle functions for the tasks assessed are either exact functions or approximate oracles developed by domain experts. Specifically, the \textbf{Branin} and \textbf{TF-Bind-8} tasks utilize exact oracles described in detail by \citet{branin} and \citet{tfbind8}, respectively. The oracle for the Penalized \textbf{LogP} task is an approximate oracle from \citet{logporacle} that is the same oracle used by domain experts in the Guacamol benchmarking study \citep{guacamol}. The \textbf{GFP}, \textbf{UTR}, and \textbf{ChEMBL} tasks feature approximate oracles from \citet{snoek2012}, \citet{utroracle}, and \citet{design-bench}, respectively, that were trained on a larger, hidden datasets inaccessible to us for the respective tasks. The \textbf{D'Kitty} morphology task uses a MuJoCo \citep{mujoco} simulation environment and learned control policy from \citet{design-bench} to evaluate proposed designs. Finally, the \textbf{Warfarin} task uses a linear model \citep{warfarin} to estimate a patient's optimal warfarin dose given their pharmacogenetic attributes. 

\textbf{Data Preprocessing.}
(1) For the \textbf{Branin} task, we sample 1000 points from the square input domain $[-5, 10]\times [0, 15]$ to construct the offline dataset, and remove the top 20\%-ile according to the oracle function to make the task more challenging in line with prior work \citep{bonet}. In this continuous task (along with the \textbf{D'Kitty} and \textbf{Warfarin} tasks), we treat input designs as their own latent space mappings, such that the VAE encoder and decoder for this task are both the identity function with zero trainable parameters. (2) The offline dataset of the Penalized \textbf{LogP} task is the validation partition of the Guacamol dataset from \citet{guacamol}, which consists of 79,564 unique molecules and their corresponding penalized LogP scores. The input molecules are represented as SMILES strings \citep{smiles}, which is a molecule representation format shown to frequently yield invalid molecules in prior work \citep{selfies}. Therefore, we encode the molecules instead as SELFIES strings, an alternative molecule representation from \citet{selfies} with 100\% robustness.

(3) - (5) The \textbf{TF-Bind-8}, \textbf{GFP}, and \textbf{UTR} tasks are assessed as-released by Design-Bench from \citet{design-bench}\textemdash please refer to their work for task-specific descriptions. (6) - (7) In the \textbf{ChEMBL} and \textbf{D'Kitty} tasks, we normalize all objective values $y$ in the offline dataset to $\hat{y}=(y-y_{\text{min}})/(y_{\text{max}}-y_{\text{min}})$ as done in prior work \citep{bonet}, where $\hat{y}$ is the corresponding normalized objective value and $y_{\text{min}}$ ($y_{\text{max}}$) is the minimum (maximum) observed objective value in the full, \textit{unobserved} dataset. Because only the bottom 60\%-ile (40\%-ile) from the full dataset is used in the available offline dataset for the ChEMBL (D'Kitty) task, the respective maximum $\hat{y}$ values are less than 1.0 (\textbf{Supplementary Table \ref{table:datasets}}). We also translate the original SMILES string representations in the ChEMBL task into SELFIES strings \citep{selfies} as in the LogP task.

(8) Finally, the \textbf{Warfarin} task uses the dataset of pharmacogenetic patient covariates published by \citet{warfarin}. We split the original dataset of 3,936 unique patient observations into training (validation) partitions with 3,736 (200) datums. The patient attributes in the Warfarin dataset consist of a combination of discrete and continuous values. All discrete attributes are one-hot encoded into binarized dimensions, and continuous values are normalized to zero mean and unit variance using the training dataset. Missing patient values were imputed following prior work \citep{warfit}. We define the cost $c(z|x)$ accrued by a patient with attributes $x\in\mathbb{R}^{32}$ as a function of the input dose $z\in\mathbb{R}$ is $c(z|x)=(z-d_{\text{oracle}}(x))^2$, where $d_{\text{oracle}}:\mathbb{R}^{32}\rightarrow \mathbb{R}$ is the domain-expert oracle warfarin dose estimator from \citet{warfarin}. The observed objective values $y$ associated with each of the training datums is calculated as $y=[c(\bar{z}|x) -c(z|x)]/c(\bar{z}|x)$, where $\bar{z}$ is the mean warfarin dose over the training dataset and $z$ is the true dose given to the patient. Using this constructed offline dataset, our task is then to assign optimal doses to the 200 validation patients to maximize $y$ with \textit{no} prior warfarin dosing observations.

\begin{table*}[htbp]
\caption{\textbf{MBO Datasets and Tasks}$\quad$ Implementation details for each of the seven MBO tasks assessed in our work. $^*$Denotes the life sciences-related discrete MBO tasks offered by the Design-Bench benchmarking repository \citep{design-bench}.\vspace{-2ex}}
\label{table:datasets}
\begin{center}
\begin{small}
\resizebox{\textwidth}{!}{\begin{tabular}{rcccccccc}
\toprule
\textbf{Property} & \textbf{Branin} & \textbf{LogP} & \textbf{TF-Bind-8$^*$} & \textbf{GFP$^*$} & \textbf{UTR$^*$} & \textbf{ChEMBL$^*$} & \textbf{D'Kitty} & \textbf{Warfarin}\\
\midrule
Dataset Size & 800 & 79,564 & 32,898 & 5,000 & 140,000 & 441 & 10,004 & 200\\
Input Shape & 2 & 108 & 8 & 237 & 50 & 32 & 56 & 1 (33)\\
Vocab Size & \textemdash & 97 & 4 & 20 & 4 & 40 & \textemdash & \textemdash\\
VAE Backbone & Identity & Transformer & ResNet & ResNet & ResNet & Transformer & Identity & Identity\\
VAE Latent Shape & 2 & 256 & 16 & 32 & 32 & 128 & 56 & 33\\
Oracle & Exact & Linear & Exact & Transformer & ResNet & Random Forest & Exact & Linear\\
$\mathcal{D}$ (best) & -13.0 & 11.3 & 0.439 & 3.53 & 7.12 & 0.61 & 0.88 & -0.19 $\pm$ 1.96\\
\bottomrule
\end{tabular}}
\end{small}
\end{center}
\end{table*}

\setcounter{table}{0}
\renewcommand{\thetable}{B\arabic{table}}
\setcounter{figure}{0}
\renewcommand{\thefigure}{B\arabic{figure}}
\section{Additional Experimental Results}
\label{appendix:b}

In this section, we provide additional experimental results that help better characterize both the strengths and limitations of GABO and GAGA.

\subsection{How do sub-optimal design candidates proposed by GABO and GAGA perform?}
\label{subsection:suboptimal}

To evaluate the robustness of optimization methods, we report one-shot 90th percentile oracle scores in \textbf{Supplementary Tables \ref{table:results-90}} and \textbf{\ref{table:ablation-90}}. For each method, all proposed designs are ranked according to the surrogate forward model ((\ref{eq:lagrangian}) for Generative Adversarial Bayesian Optimization (GABO) and Generative Adversarial Gradient Ascent (GAGA)), and the single 90th percentile design according to this ranking is selected and evaluated using the oracle function. We report the oracle score of this suboptimal design averaged over 10 seeds.

Our results show that GABO and GAGA do not propose suboptimal designs that are better than those proposed by other methods, such as BONET \citep{bonet}, Simulated Annealing \citep{simanneal}, L-BFGS \citep{lbfgs}, and ExPT \citep{expt}. This is not surprising, as aSCR is not designed to target this metric (and it is not our primary metric of interest). Separately for GABO, we also hypothesize that the algorithm's performance according to this metric may partially be explained by the limitations of the underlying Bayesian optimization (BO) optimization algorithm. Because BO is not an iterative first-order algorithm, the designs proposed by any BO-based algorithm often have high variance in practice\textemdash this is indeed what we observe across all of our experiments, including in \textbf{Table \ref{table:results-top-1}} and \textbf{Supplementary Tables \ref{table:results-90}} and \textbf{\ref{table:ablation-90}}.

Finally, we note that in most applications of offline optimization, the 90th percentile metric\textemdash or any metric that does not use the best proposed design(s)\textemdash is not as useful as the other metrics assessed where GABO does perform well. This is because in offline optimization tasks with a restricted budget to query the hidden, expensive-to-evaluate oracle function, we are not interested in ``wasting'' this limited budget on subpar design candidates. While the 90th percentile and similar metrics can be helpful to understand the limitations of algorithms (as in this case), we believe that the alternative evaluation metrics reported in the main text\textemdash namely, the 100th percentile top-1 and top-128 oracle score metrics\textemdash are more useful and practical in assessing each of the optimization algorithms.

\begin{table*}[ht]
\caption{\textbf{Constrained Budget ($k=1$) Suboptimal (90\%-ile) Oracle Evaluation}$\quad$ The oracle score of the 90th percentile design candidate according to the surrogate across 10 random seeds is reported as mean $\pm$ standard deviation. $\mathcal{D}$ (best) reports the top oracle value in the task dataset. The average rank across all seven tasks is reported in the final table column. \textbf{Bolded} (\underline{Underlined}) entries indicate the best (second best) entry in the column. $^*$Denotes the life sciences-related discrete MBO tasks from Design-Bench \citep{design-bench}.}
\label{table:results-90}
\begin{center}
\begin{small}
\resizebox{\textwidth}{!}{\begin{tabular}{rccccccccc}
\toprule
\textbf{Method} & \textbf{Branin} & \textbf{LogP} & \textbf{TF-Bind-8$^*$} & \textbf{GFP$^*$} & \textbf{UTR$^*$} & \textbf{ChEMBL$^*$} & \textbf{D'Kitty} & \textbf{Warfarin} & \textbf{Rank}\\
\midrule
$\mathcal{D}$ (best) & -13.0 & 11.3 & 0.439 & 3.53 & 7.12 & 0.61 & 0.88 & -0.19 $\pm$ 1.96 & --- \\
\midrule
Grad. & -94.4 $\pm$ 20.9 & -5.47 $\pm$ 1.32 & 0.429 $\pm$ 0.023 & 3.43 $\pm$ 0.67 & 7.16 $\pm$ 0.21 & -1.95 $\pm$ 0.00 & 0.53 $\pm$ 0.20 & \underline{0.87 $\pm$ 1.08} & 9.1 \\
L-BFGS & \textbf{-4.0 $\pm$ 0.0} & 4.96 $\pm$ 6.64 & 0.547 $\pm$ 0.163 & 3.50 $\pm$ 0.70 & 7.36 $\pm$ 0.92 & -1.95 $\pm$ 0.00 & 0.31 $\pm$ 0.00 & 0.75 $\pm$ 1.66 & \underline{6.9} \\
CMA-ES & \underline{-10.4 $\pm$ 3.0} & -4.35 $\pm$ 6.18 & 0.448 $\pm$ 0.068 & \textbf{3.74 $\pm$ 0.00} & 6.95 $\pm$ 1.13 & -1.95 $\pm$ 0.00 & 0.60 $\pm$ 0.29 & -4.02 $\pm$ 21.8 & 7.1 \\
Anneal & -13.2 $\pm$ 0.0 & \underline{9.57 $\pm$ 0.66} & 0.439 $\pm$ 0.000 & \underline{3.65 $\pm$ 0.04} & 7.41 $\pm$ 0.22 & -1.95 $\pm$ 0.00 & 0.56 $\pm$ 0.00 & \textbf{0.96 $\pm$ 0.08} & \textbf{6.0} \\
BO & -11.5 $\pm$ 2.3 & -56.2 $\pm$ 91.9 & \underline{0.552 $\pm$ 0.152} & 1.42 $\pm$ 0.00 & 5.80 $\pm$ 1.71 & \underline{0.64 $\pm$ 0.01} & 0.46 $\pm$ 0.18 & -36.9 $\pm$ 205 & 9.6 \\
TuRBO & -16.3 $\pm$ 10.2 & -24.3 $\pm$ 66.3 & \textbf{0.563 $\pm$ 0.087} & 1.42 $\pm$ 0.00 & 6.79 $\pm$ 1.25 & \textbf{0.65 $\pm$ 0.00} & 0.71 $\pm$ 0.01 & -32.3 $\pm$ 94.9 & 7.9 \\
BONET & -29.2 $\pm$ 2.2 & \textbf{10.8 $\pm$ 0.43} & 0.324 $\pm$ 0.041 & \textbf{3.74 $\pm$ 0.00} & \textbf{8.70 $\pm$ 0.32} & 0.56 $\pm$ 0.11 & 0.78 $\pm$ 0.00 & --- & \textbf{6.0} \\
DDOM & -1870 $\pm$ 2693 & -7.10 $\pm$ 1.42 & 0.386 $\pm$ 0.224 & 1.43 $\pm$ 0.00 & \underline{7.91 $\pm$ 0.29} & \textbf{0.65 $\pm$ 0.01} & 0.50 $\pm$ 0.19 & -56.6 $\pm$ 79.6 & 9.6 \\
COM & -3468 $\pm$ 679 & -37.4 $\pm$ 23.0 & 0.346 $\pm$ 0.093 & 3.62 $\pm$ 0.00 & 5.26 $\pm$ 1.01 & 0.60 $\pm$ 0.04 & \textbf{0.90 $\pm$ 0.01} & 0.80 $\pm$ 0.93 & 9.6 \\
RoMA & -18.5 $\pm$ 8.2 & 5.21 $\pm$ 1.39 & 0.500 $\pm$ 0.153 & 3.58 $\pm$ 0.11 & 6.94 $\pm$ 1.11 & 0.43 $\pm$ 0.18 & 0.41 $\pm$ 0.21 & -2.44 $\pm$ 2.16 & 8.1 \\
BDI & -109 $\pm$ 0.0 & 0.93 $\pm$ 0.88 & 0.471 $\pm$ 0.000 & 3.58 $\pm$ 0.05 & 5.62 $\pm$ 0.00 & 0.49 $\pm$ 0.00 & 0.76 $\pm$ 0.00 & -24.8 $\pm$ 233 & 9.0 \\
ExPT & -23.1 $\pm$ 11.3 & -16.7 $\pm$ 25.1 & 0.480 $\pm$ 0.091 & \textbf{3.74 $\pm$ 0.00} & 6.70 $\pm$ 0.39 & 0.62 $\pm$ 0.04 & 0.75 $\pm$ 0.07 & -0.40 $\pm$ 1.61 & \underline{6.9} \\
BootGen & --- & -116.8 $\pm$ 85.7 & 0.388 $\pm$ 0.007 & 3.60 $\pm$ 0.04 & 7.74 $\pm$ 0.56 & 0.61 $\pm$ 0.03 & --- & --- & 8.8 \\
ROMO & -3142 $\pm$ 330 & -25.6 $\pm$ 23.1 & 0.354 $\pm$ 0.247 & 3.59 $\pm$ 0.08 & 5.49 $\pm$ 1.38 & 0.62 $\pm$ 0.04 & 0.42 $\pm$ 0.17 & -2.77 $\pm$ 5.21 & 11.1 \\
\midrule
\textbf{GAGA} & -14.2 $\pm$ 15.2 & -16.7 $\pm$ 81.1 & 0.546 $\pm$ 0.148 & 3.22 $\pm$ 0.86 & 6.40 $\pm$ 1.13 & -1.95 $\pm$ 0.00 & \underline{0.89 $\pm$ 0.01} & 0.24 $\pm$ 0.20 & 8.5 \\
\textbf{GABO} & -12.7 $\pm$ 10.0 & -12.2 $\pm$ 46.1 & 0.467 $\pm$ 0.066 & 3.56 $\pm$ 1.66 & 6.12 $\pm$ 1.22 & 0.61 $\pm$ 0.08 & 0.57 $\pm$ 0.17 & 0.02 $\pm$ 5.77 & 7.9 \\
\bottomrule
\end{tabular}}
\end{small}
\end{center}
\end{table*}

\begin{table*}[ht]
\vskip -0.1in
\caption{\textbf{GABO Adaptive SCR Ablation Study\textemdash Constrained Budget ($k=1$) Suboptimal (90\%-ile) Oracle Evaluation}$\quad$ The oracle score of the 90th percentile design candidate according to the surrogate across 10 random seeds reported as mean $\pm$ standard deviation. $\mathcal{D}$ (best) reports the top oracle value in the task dataset. Average method rank across all seven tasks reported in the final column. $^*$Denotes the life sciences-related discrete MBO tasks Design-Bench \citep{design-bench}.}
\label{table:ablation-90}
\begin{center}
\begin{small}
\resizebox{\textwidth}{!}{\begin{tabular}{rcccccccccc}
\toprule
\textbf{GABO $\alpha$ Value} & \textbf{Branin} & \textbf{LogP} & \textbf{TF-Bind-8$^*$} & \textbf{GFP$^*$} & \textbf{UTR$^*$} & \textbf{ChEMBL$^*$} & \textbf{D'Kitty} &  \textbf{Warfarin} & \textbf{Rank}\\
\midrule
$\mathcal{D}$ (best) & -13.0 & 11.3 & 0.439 & 3.53 & 7.12 & 0.61 & 0.88 & -0.19 $\pm$ 1.96 & --- \\
\midrule
$\alpha=0.0$ & -11.5 $\pm$ 2.3 & -56.2 $\pm$ 91.9 & \underline{0.552 $\pm$ 0.152} & 1.42 $\pm$ 0.00 & 5.80 $\pm$ 1.71 & \textbf{0.64 $\pm$ 0.01} & 0.46 $\pm$ 0.18 & \underline{-36.9 $\pm$ 205} & 3.3 \\
$\alpha=0.2$ & \underline{-9.0 $\pm$ 2.6} & \underline{-40.2 $\pm$ 77.4} & \textbf{0.612 $\pm$ 0.114} & 1.42 $\pm$ 0.00 & 5.81 $\pm$ 1.83 & 0.59 $\pm$ 0.13 & \underline{0.49 $\pm$ 0.18} & -51.7 $\pm$ 265 & \underline{2.9} \\
$\alpha=0.5$ & \textbf{-8.6 $\pm$ 4.4} & -90.1 $\pm$ 107.2 & 0.501 $\pm$ 0.109 & 1.65 $\pm$ 0.69 & \textbf{6.64 $\pm$ 1.42} & 0.52 $\pm$ 0.15 & 0.41 $\pm$ 0.16 & -63.5 $\pm$ 336 & 3.9 \\
$\alpha=0.8$ & -10.9 $\pm$ 2.1 & -41.9 $\pm$ 82.5 & 0.433 $\pm$ 0.158 & 1.97 $\pm$ 0.88 & 4.89 $\pm$ 1.23 & 0.56 $\pm$ 0.15 & 0.38 $\pm$ 0.15 & -48.5 $\pm$ 265 & 4.4 \\
$\alpha=1.0$ & -104.6 $\pm$ 68.9 & -77.1 $\pm$ 146.1 & 0.452 $\pm$ 0.179 & \underline{2.05 $\pm$ 0.98} & 5.15 $\pm$ 1.51 & 0.60 $\pm$ 0.08 & 0.41 $\pm$ 0.16 & -82.1 $\pm$ 552 & 4.5 \\
\midrule
\textbf{aSCR} & -12.7 $\pm$ 10.0 & \textbf{-12.2 $\pm$ 46.1} & 0.467 $\pm$ 0.066 & \textbf{3.56 $\pm$ 1.66} & \underline{6.12 $\pm$ 1.22} & \underline{0.61 $\pm$ 0.08} & \textbf{0.57 $\pm$ 0.17} & \textbf{0.02 $\pm$ 5.77} & \textbf{2.1} \\
\bottomrule
\end{tabular}}
\end{small}
\end{center}
\vskip -0.1in
\end{table*}

\begin{figure}
    \centering
    \includegraphics[width=0.89\linewidth]{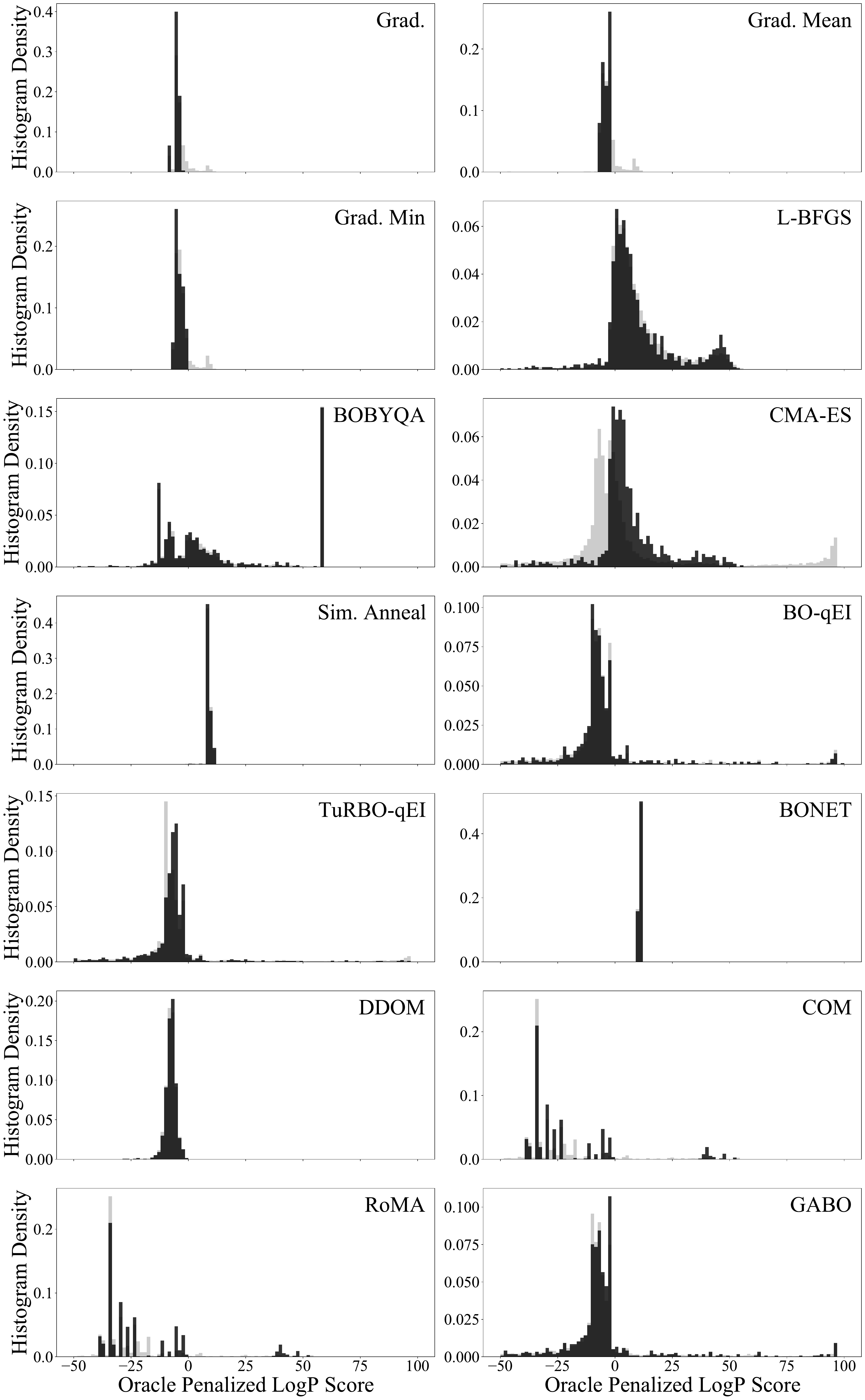}
    \caption{\textbf{Distribution of Oracle Penalized LogP Scores}\quad We plot the distribution of oracle scores for the top 128 surrogate model-ranked designs in black, and the distribution for all 2,048 generated designs in light gray for each of the offline model-based optimization methods assessed in our work across 10 random seeds. While GABO and BO-qEI have similar distributions, GABO is able to more reliably rank top-performing designs higher, such that these designs can be identified even under limited oracle query budgets.}
    \label{fig:distribution}
\end{figure}

To further characterize the distribution of designs and their associated oracle scores proposed by GABO, \textbf{Figure \ref{fig:distribution}} plots a histogram of the oracle scores of (1) all 2,048 oracle scores, and (2) the oracle scores of the top 256 designs according to the penalized surrogate objective in (\ref{eq:lagrangian}) for the \textbf{LogP} task. Compared with the other optimization methods assessed, we notice that the range of oracle scores is larger for BO-based optimization methods compared with the baseline methods assessed. This helps motivate our design choice to leverage aSCR and \textbf{Algorithm \ref{alg:ascr}} with BO-qEI, as BO is able to explore a larger region of the design space and is an effective parent optimizer for complex design spaces. Secondly, we also find that the distribution of scores is similar between BO-qEI and GABO, even though the performance of these two methods is remarkably different in \textbf{Tables \ref{table:results-top-1}} and \textbf{\ref{table:results-top-128}}. This is likely due to the fact that while BO enables us to explore a larger effective region of the design space (compared with first-order iterative methods), \textbf{aSCR more accurately ranks proposed designs using the penalized surrogate so that we can identify promising candidates even in the low-budget oracle evaluation regime}.

\subsection{Are offline objectives and oracle function values correlated?}
A key component of GABO with Adapative SCR critical to the above discussion in \textbf{Section \ref{subsection:suboptimal}} is that generated designs score similarly according to the hidden oracle function and the regularized Lagrangian objective as in (\ref{eq:lagrangian}) in order to solve the problem of surrogate objective overestimation encountered in traditional offline optimization settings (\textbf{Fig. \ref{fig:overview}}). To assess this quantitatively, we computed the distance covariance $\text{dCov}_n[\{\mathcal{L}(\mathbf{x}_k; \lambda^*)\}_{k=1}^n, \{f(\mathbf{x}_k)\}_{k=1}^n]$ between the oracle scores $f(\mathbf{x}_k)$ and the constrained Lagrangian scores $\mathcal{L}(\mathbf{x}_k; \lambda^*)$ with $\lambda=\lambda^*(t)$ computed using our Adaptive SCR algorithm. The empirical distance covariance metric is computed over the $n=2048$ design candidates generated using our GABO algorithm. Briefly, the distance covariance is a nonnegative measure of dependence between two vectors which may be related nonlinearly; a greater distance covariance implies a greater degree of association between observations \citep{distcovar}. We focus our subsequent discussion on the Penalized \textbf{LogP} task.

Across five random seeds, GABO with Adaptive SCR achieves a distance covariance score of 0.535 $\pm$ 0.067 (mean $\pm$ standard deviation). In contrast, na\"{i}ve BO-qEI (i.e., $\lambda=0$) only achieves a distance covariance score of 0.392 $\pm$ 0.040. Using $p<0.05$ as a cutoff for statistical significance, the distance covariance scores are significantly different between these two methods ($p\approx 0.006$, unpaired two-tailed $t$-test). These results help support our conclusion that GABO with Adaptive SCR is able to provide better estimates of design candidate performance according to the hidden oracle function when compared to the corresponding unconstrained BO policy.

\subsection{Is adaptively computing \texorpdfstring{$\alpha$}{alpha} in aSCR important for the performance of GAGA?}

In our ablation experiments presented in \textbf{Table \ref{table:ablation}}, we showed how that `adaptive' nature of aSCR is an important component in solving the constrained optimization problem in (\ref{eq:constrained-opt}) for GABO, and outperforms alternative approaches that manually hand-tune $\alpha$ (and hence $\lambda$) as a constant hyperparameter. We explore whether this conclusion also applies for GAGA as well here.

For clarity, we first offer the explicit formulation of GAGA in \textbf{Supplementary Algorithm \ref{alg:gaga}}. We ablate \textbf{Algorithm \ref{alg:ascr}} in GAGA by instead evaluating our method using different values of $\lambda=\alpha/(1-\alpha)$. As a reminder, setting $\alpha=0$ (i.e., $\lambda=0$) corresponds to na\"{i}vely performing gradient ascent against the unconstrained surrogate model; setting $\alpha=1$ (i.e., $\lambda\rightarrow \infty$) is equivalent to a WGAN-like generative policy.

\begin{algorithm}[H]
  \caption{\small Generative Adversarial Gradient Ascent (GAGA)}
  \label{alg:gaga}
  \small\begin{algorithmic}
    \STATE {\bfseries Input:} surrogate objective $f_{\theta}: \mathbb{R}^d\rightarrow \mathbb{R}$, offline dataset $\mathcal{D}_n=\{z'_j\}_{j=1}^n$, iterative sampling budget $T$, sampling batch size $b$, number of generator steps per source critic training $n_{\text{generator}}$, oracle query budget $k$, step size $\eta$
    \STATE {\bfseries AdaptiveSCR Input:} $\alpha$ step size $\Delta \alpha$, search budget $\mathcal{B}$, norm threshold $\tau$
    \STATE \textbf{Define:} Differentiable source critic $c: \mathbb{R}^d\rightarrow \mathbb{R}$
    \STATE {\bfseries Define:} Lagrangian $\mathcal{L}(z; \alpha):\mathbb{R}^d\times \mathbb{R}\rightarrow \mathbb{R}=-f_{\theta}(z)+\frac{\alpha}{1-\alpha}[\mathbb{E}_{z'\sim\mathcal{D}_n}[c(z')]-c(z)]$ \hspace{3ex} // Eq. (\ref{eq:lagrangian})
    \STATE Sample $\mathcal{Z}^1\leftarrow \{z^1_i\}_{i=1}^b$ as the top $b$ designs in $\mathcal{D}_n$ according to their previously observed oracle scores
    \STATE // Train the source critic per Eq. (\ref{eq:kr-duality}) to optimality:
    \STATE $c \leftarrow \text{argmax}_{||c||_L\leq K} W_1(\mathcal{D}_n, \mathcal{Z}^1)=\text{argmax}_{||c||_L\leq K}\left[\mathbb{E}_{z'\sim \mathcal{D}_n}[c(z')] - \mathbb{E}_{z\sim \mathcal{Z}^1}[c(z)]\right]$
    \STATE $\alpha\leftarrow \textbf{AdaptiveSCR}(f_{\theta}, c, \mathcal{D}_n, \Delta\alpha, \mathcal{B}, \tau)$ \hspace{4ex}// Alg. (\ref{alg:ascr})
    \STATE Evaluate candidates $\mathcal{Y}^1\leftarrow \{y^1_i\}_{i=1}^b=\{-\mathcal{L}(z^1_i; \alpha)\}_{i=1}^b$
    \FOR{$t$ {\bfseries in} $2, 3, \ldots, T$}
      \STATE $\mathcal{Z}^{t}\leftarrow \{z^t_i\}_{i=1}^b = \{z^{t-1}_i-\eta\nabla_{z^{t-1}_i}\mathcal{L}(z^{t-1}_i; \alpha)\}_{i=1}^b$
      \STATE $\alpha\leftarrow \textbf{AdaptiveSCR}(f_{\theta}, c, \mathcal{D}_n, \Delta\alpha, \mathcal{B}, \tau)$
      \STATE Evaluate samples $\mathcal{Y}^t\leftarrow \{y^t_i\}_{i=1}^b=\{-\mathcal{L}(z_i^t; \alpha)\}_{i=1}^b$
      \IF{$t \text{ mod } n_{\text{generator}}$ equals $0$}
        \STATE // Train the source critic per Eq. (\ref{eq:kr-duality}) to optimality:
        \STATE $c \leftarrow \text{argmax}_{||c||_L\leq K} W_1(\mathcal{D}_n, \mathcal{Z}^t)=\text{argmax}_{||c||_L\leq K}\left[\mathbb{E}_{z'\sim \mathcal{D}_n}[c(z')] - \mathbb{E}_{z\sim \mathcal{Z}^t}[c(z)]\right]$
      \ENDIF
    \ENDFOR
    \STATE \textbf{return} the top $k$ samples from the $T\times b$ observations $\mathcal{D}_T=\{\{(z^m_i, y^m_i)\}_{i=1}^b\}_{m=1}^T$ according to $y^m_i$
  \end{algorithmic}
\end{algorithm}

Our results are shown in \textbf{Supplementary Table \ref{table:gaga-ablation}}: similar to the analogous ablation results for GABO in \textbf{Table \ref{table:ablation}}, dynamically adjusting the strength of source critic regularization using our aSCR algorithm outperforms manually setting the value of $\alpha$ to a constant in both the one-shot $k=1$ and few-shot $k=128$ evaluation settings. 

\begin{table*}[t]
\caption{\textbf{GAGA Adaptive ACR Ablation Study}$\quad$ We ablate the dynamic computation of $\alpha$ (and hence $\lambda$ in (\ref{eq:lagrangian})) by instead choosing to manually fix $\alpha$ to a constant value. A value of $\alpha=0.0$ corresponds to na\"{i}ve gradient ascent, and a value of $\alpha=1.0$ corresponds to a WGAN-like generative policy. Oracle values are averaged across 10 random seeds and reported as mean $\pm$ standard deviation. In each evaluation setting, we rank all 2,048 proposed designs according to the penalized surrogate forward model in (\ref{eq:lagrangian}) and evaluate the top $k$ designs using the oracle function, reporting the maximum out of the $k$ oracle values. In the suboptimal evaluation setting, we report the oracle score of the single 90th percentile design according to the penalized surrogate ranking. \textbf{Bold} (\underline{Underlined}) entries indicate the best (second best) entry in the column for the particular evaluation metric. $^*$Denotes the life sciences MBO tasks offered by Design-Bench \citep{design-bench}.}
\label{table:gaga-ablation}
\begin{center}
\begin{small}
\resizebox{\textwidth}{!}{\begin{tabular}{rccccccccc}
\toprule
 & \textbf{Branin} & \textbf{LogP} & \textbf{TF-Bind-8$^*$} & \textbf{GFP$^*$} & \textbf{UTR$^*$} & \textbf{ChEMBL$^*$} & \textbf{D'Kitty} & \textbf{Warfarin} & \textbf{Rank}\\
\midrule
$\mathcal{D}$ (best) & -13.0 & 11.3 & 0.439 & 3.53 & 7.12 & 0.61 & 0.88 & -0.19 $\pm$ 1.96 & ---\\
\midrule
\multicolumn{10}{c}{\textbf{Constrained Budget $(k=1)$ Oracle Evaluation}}\\
\midrule
$\alpha=0.0$ & -245.1 $\pm$ 81.3 & \textbf{-5.37 $\pm$ 1.44} & 0.429 $\pm$ 0.023 & \underline{3.18 $\pm$ 0.88} & \underline{6.82 $\pm$ 0.21} & -1.95 $\pm$ 0.00 & 0.57 $\pm$ 0.19 & \textbf{0.86 $\pm$ 1.09} & 3.6 \\
$\alpha=0.2$ & \underline{-13.7 $\pm$ 0.0} & -70.3 $\pm$ 115.8 & \underline{0.439 $\pm$ 0.000} & \textbf{3.74 $\pm$ 0.00} & \textbf{7.73 $\pm$ 0.46} & -1.95 $\pm$ 0.00 & \underline{0.88 $\pm$ 0.00} & -0.17 $\pm$ 0.00 & 2.5 \\
$\alpha=0.5$ & \underline{-13.7 $\pm$ 0.0} & -70.3 $\pm$ 114.7 & \underline{0.439 $\pm$ 0.000} & \textbf{3.74 $\pm$ 0.00} & 6.75 $\pm$ 0.72 & -1.95 $\pm$ 0.00 & \underline{0.88 $\pm$ 0.00} & \underline{0.44 $\pm$ 0.00} & \underline{2.4} \\
$\alpha=0.8$ & \underline{-13.7 $\pm$ 0.0} & -84.6 $\pm$ 115.8 & \underline{0.439 $\pm$ 0.000} & \textbf{3.74 $\pm$ 0.00} & 6.75 $\pm$ 0.72 & -1.95 $\pm$ 0.00 & \underline{0.88 $\pm$ 0.00} & \underline{0.44 $\pm$ 0.00} & 2.6 \\
$\alpha=1.0$ & -14.4 $\pm$ 1.5 & \underline{-27.8 $\pm$ 99.8} & \underline{0.439 $\pm$ 0.000} & \textbf{3.74 $\pm$ 0.00} & 5.88 $\pm$ 1.04 & -1.95 $\pm$ 0.00 & \textbf{0.89 $\pm$ 0.00} & -8.61 $\pm$ 6.15 & 3.4 \\
\midrule
\textbf{aSCR} & \textbf{-2.9 $\pm$ 2.2} & -68.6 $\pm$ 109.8 & \textbf{0.571 $\pm$ 0.120} & \textbf{3.74 $\pm$ 0.00} & 5.89 $\pm$ 1.42 & -1.95 $\pm$ 0.00 & \textbf{0.89 $\pm$ 0.00} & 0.01 $\pm$ 0.14 & \textbf{2.3} \\
\bottomrule
\\
\toprule
\multicolumn{10}{c}{\textbf{Relaxed Budget $(k=128)$ Oracle Evaluation}}\\
\midrule
$\alpha=0.0$ & -115.3 $\pm$ 20.8 & -5.14 $\pm$ 1.70 & \textbf{0.977 $\pm$ 0.025} & \underline{3.49 $\pm$ 0.69} & 7.38 $\pm$ 0.15 & -1.95 $\pm$ 0.00 & 0.87 $\pm$ 0.02 & 0.86 $\pm$ 1.08 & 4.8 \\
$\alpha=0.2$ & -13.2 $\pm$ 0.0 & 4.70 $\pm$ 10.3 & 0.439 $\pm$ 0.000 & \textbf{3.74 $\pm$ 0.00} & \underline{7.92 $\pm$ 0.24} & -1.95 $\pm$ 0.00 & \textbf{0.95 $\pm$ 0.00} & \textbf{1.00 $\pm$ 0.00} & \underline{2.4} \\
$\alpha=0.5$ & -13.2 $\pm$ 0.0 & 5.07 $\pm$ 4.56 & 0.439 $\pm$ 0.000 & \textbf{3.74 $\pm$ 0.00} & 7.77 $\pm$ 0.21 & -1.95 $\pm$ 0.00 & \textbf{0.95 $\pm$ 0.01} & \textbf{1.00 $\pm$ 0.00} & 2.9 \\
$\alpha=0.8$ & -13.2 $\pm$ 0.0 & \underline{5.13 $\pm$ 4.28} & 0.439 $\pm$ 0.000 & \textbf{3.74 $\pm$ 0.00} & 7.44 $\pm$ 0.30 & -1.95 $\pm$ 0.00 & \textbf{0.95 $\pm$ 0.01} & \textbf{1.00 $\pm$ 0.00} & 2.9 \\
$\alpha=1.0$ & \underline{-13.1 $\pm$ 0.0} & 5.11 $\pm$ 4.11 & 0.445 $\pm$ 0.017 & \textbf{3.74 $\pm$ 0.00} & 7.40 $\pm$ 0.28 & -1.95 $\pm$ 0.00 & \underline{0.90 $\pm$ 0.01} & \underline{0.96 $\pm$ 0.05} & 3.0 \\
\midrule
\textbf{aSCR} & \textbf{-1.0 $\pm$ 0.2} & \textbf{14.1 $\pm$ 25.0} & \underline{0.722 $\pm$ 0.091} & \textbf{3.74 $\pm$ 0.00} & \textbf{7.98 $\pm$ 0.36} & -1.95 $\pm$ 0.00 & \underline{0.90 $\pm$ 0.01} & 0.95 $\pm$ 0.07 & \textbf{2.1} \\
\bottomrule
\\
\toprule
\multicolumn{10}{c}{\textbf{Constrained Budget $(k=1)$ Suboptimal (90\%-ile) Oracle Evaluation}}\\
\midrule
$\alpha=0.0$ & -94.4 $\pm$ 20.9 & \textbf{-5.47 $\pm$ 1.32} & 0.429 $\pm$ 0.023 & \underline{3.43 $\pm$ 0.67} & \textbf{7.16 $\pm$ 0.21} & -1.95 $\pm$ 0.00 & 0.53 $\pm$ 0.20 & 0.87 $\pm$ 1.08 & 3.6 \\
$\alpha=0.2$ & -18.1 $\pm$ 0.5 & \underline{-10.9 $\pm$ 14.9} & 0.439 $\pm$ 0.000 & \textbf{3.74 $\pm$ 0.00} & 6.57 $\pm$ 0.94 & -1.95 $\pm$ 0.00 & \underline{0.89 $\pm$ 0.02} & \textbf{0.97 $\pm$ 0.04} & \underline{2.5} \\
$\alpha=0.5$ & -16.2 $\pm$ 0.6 & -15.2 $\pm$ 14.4 & \underline{0.445 $\pm$ 0.017} & \textbf{3.74 $\pm$ 0.00} & 6.75 $\pm$ 1.18 & -1.95 $\pm$ 0.00 & \textbf{0.90 $\pm$ 0.02} & \underline{0.93 $\pm$ 0.18} & \textbf{2.4} \\
$\alpha=0.8$ & -15.7 $\pm$ 1.0 & -12.7 $\pm$ 13.8 & 0.439 $\pm$ 0.000 & \textbf{3.74 $\pm$ 0.00} & \underline{6.84 $\pm$ 1.29} & -1.95 $\pm$ 0.00 & 0.88 $\pm$ 0.01 & -0.24 $\pm$ 2.89 & 3.1 \\
$\alpha=1.0$ & \underline{-14.6 $\pm$ 1.4} & -16.9 $\pm$ 13.1 & 0.439 $\pm$ 0.000 & \textbf{3.74 $\pm$ 0.00} & 6.82 $\pm$ 1.01 & -1.95 $\pm$ 0.00 & \underline{0.89 $\pm$ 0.01} & -2.71 $\pm$ 7.71 & 3.3 \\
\midrule
\textbf{aSCR} & \textbf{-14.2 $\pm$ 15.2} & -16.7 $\pm$ 81.1 & \textbf{0.546 $\pm$ 0.148} & 3.22 $\pm$ 0.86 & 6.40 $\pm$ 1.13 & -1.95 $\pm$ 0.00 & \underline{0.89 $\pm$ 0.01} & 0.24 $\pm$ 0.20 & 3.5 \\
\bottomrule
\end{tabular}}
\end{small}
\end{center}
\end{table*} 

\subsection{What is the impact of dynamic updates to the source critic over the optimization trajectory?}
In \textbf{Algorithm \ref{alg:gabo}} and \textbf{Supplementary Algorithm \ref{alg:gaga}}, we describe how generative adversarial optimization alternates between batched acquisition steps according to the optimizer and re-training the source critic on the newly sampled trajectory points. To better interrogate the significance of dynamically re-training the source critic during optimization, we compare the performance of the default GABO and GAGA algorithms (with $n_{\text{generator}}=4$ as the number of acquisition steps per critic retraining step) against the respective methods without source critic re-training (i.e., $n_{\text{generator}}=\infty$) in \textbf{Supplementary Table \ref{table:ablate-online-source-critic-training}}. Across all three evaluation metrics and all eight tasks, dynamically retraining the source-critic improves upon the performance of the GABO when $n_{\text{generator}}=\infty$ by 67.4\% in the top-1 evaluation metric; 0.0\% in the top-128 evaluation metric; and 33.5\% in the 90\%-ile evaluation metric. Intuitively, these results align with the value of the source critic in being able to implicitly set the value of the regularization strength $\alpha$ in (\ref{eq:lagrangian}) according to the sampled trajectory points\textemdash especially in the constrained budget oracle evaluation setting.

Interestingly, we do not observe similar performance improvements with dynamic re-training of the source critic in GAGA. Qualitatively, we find that this is because of the iterative first-order nature of the parent gradient ascent algorithm\textemdash because the sampled designs are clustered in the same regions of the design space over the course of optimization, the energy landscape of the penalized surrogate (i.e., the negative of the Lagrangian expression in (\ref{eq:lagrangian})) does not change significantly during source critic re-training. This further reinforces the optimizer to stay roughly in the same regions of the design space. As a result, it is likely that no major updates are often made to the source critic when aSCR is used in conjunction with a first-order optimization method, and so the benefit of using a finite $n_{\text{generator}}$ hyperparameter value is largely reduced when compared to its utility in GABO.

\begin{table*}[htbp]
\caption{\textbf{Ablating Dynamic Updates to the Source Critic}$\quad$ We study the effect of training the source critic model \textit{exactly once} (i.e., setting $n_{\text{generator}}=\infty$ in \textbf{Algorithm \ref{alg:gabo}} and \textbf{Supplementary Algorithm \ref{alg:gaga}}) as opposed to re-training the source critic model every $n_{\text{generator}}=4$ acquisition steps on the newly sampled designs. Oracle values are averaged across 10 random seeds and reported as mean $\pm$ standard deviation. In each evaluation setting, we rank all 2,048 proposed designs according to the penalized surrogate forward model in (\ref{eq:lagrangian}) and evaluate the top $k$ designs using the oracle function, reporting the maximum out of the $k$ oracle values. In the suboptimal evaluation setting, we report the oracle score of the single 90th percentile design according to the penalized surrogate ranking. \textbf{Bold} entries indicate the best entry in the column for the particular optimizer and evaluation metric. $^*$Denotes the life sciences MBO tasks offered by Design-Bench \citep{design-bench}.\vspace{-2ex}}
\label{table:ablate-online-source-critic-training}
\begin{center}
\begin{small}
\resizebox{\textwidth}{!}{\begin{tabular}{rccccccccc}
\toprule
\textbf{GABO} & \textbf{Branin} & \textbf{LogP} & \textbf{TF-Bind-8$^*$} & \textbf{GFP$^*$} & \textbf{UTR$^*$} & \textbf{ChEMBL$^*$} & \textbf{D'Kitty} & \textbf{Warfarin}\\
\midrule
$\mathcal{D}$ (best) & -13.0 & 11.3 & 0.439 & 3.53 & 7.12 & 0.61 & 0.88 & -0.19 $\pm$ 1.96\\
\midrule
\multicolumn{9}{c}{\textbf{Constrained Budget $(k=1)$ Oracle Evaluation}}\\
\midrule
$n_{\text{generator}}=\infty$ & -3.5 $\pm$ 2.5 & -55.6 $\pm$ 52.1 & \textbf{0.577 $\pm$ 0.151} & \textbf{3.74 $\pm$ 0.00} & 6.73 $\pm$ 1.10 & \textbf{0.65 $\pm$ 0.00} & 0.46 $\pm$ 0.18 & -0.27 $\pm$	13.7\\
$n_{\text{generator}}=4$ & \textbf{-2.6 $\pm$ 1.1} & \textbf{21.3 $\pm$ 33.2} & 0.570 $\pm$ 0.131 & 3.60 $\pm$ 0.40 & \textbf{7.51 $\pm$ 0.39} & 0.60 $\pm$ 0.07 & \textbf{0.71 $\pm$ 0.01} & \textbf{0.60 $\pm$ 1.80}\\
\bottomrule
\toprule
\multicolumn{9}{c}{\textbf{Relaxed Budget $(k=128)$ Oracle Evaluation}}\\
\midrule
$n_{\text{generator}}=\infty$ & -0.5 $\pm$ 0.1 & \textbf{128.0 $\pm$ 19.5} & 0.946 $\pm$ 0.035 & 3.74 $\pm$ 0.00 & \textbf{8.38 $\pm$ 0.11} & 0.67 $\pm$ 0.01 & 0.72 $\pm$ 0.00 & 1.00 $\pm$ 0.00\\
$n_{\text{generator}}=4$ & -0.5 $\pm$ 0.1 & 122.1 $\pm$ 20.6 & \textbf{0.954 $\pm$ 0.025} & 3.74 $\pm$ 0.00 & 8.36 $\pm$ 0.08 & \textbf{0.70 $\pm$ 0.01} & 0.72 $\pm$ 0.00 & 1.00 $\pm$ 0.03\\
\bottomrule
\toprule
\multicolumn{9}{c}{\textbf{Constrained Budget $(k=1)$ Suboptimal (90\%-ile) Oracle Evaluation}}\\
\midrule
$n_{\text{generator}}=\infty$ & \textbf{-8.9 $\pm$ 6.6} & -54.1 $\pm$ 62.6 & \textbf{0.471 $\pm$ 0.061} & 3.06 $\pm$ 1.04 & 6.02 $\pm$ 1.41 & \textbf{0.63 $\pm$ 0.07} & 0.26 $\pm$ 0.62 & -5.32 $\pm$ 4.59\\
$n_{\text{generator}}=4$ & -12.7 $\pm$ 10.0 & \textbf{-12.2 $\pm$ 46.1} & 0.467 $\pm$ 0.066 & \textbf{3.56 $\pm$ 1.66} & \textbf{6.12 $\pm$ 1.22} & 0.61 $\pm$ 0.08 & \textbf{0.57 $\pm$ 0.17} & \textbf{0.02 $\pm$ 5.77}\\
\bottomrule
\\
\toprule
\textbf{GAGA} & \textbf{Branin} & \textbf{LogP} & \textbf{TF-Bind-8$^*$} & \textbf{GFP$^*$} & \textbf{UTR$^*$} & \textbf{ChEMBL$^*$} & \textbf{D'Kitty} & \textbf{Warfarin}\\
\midrule
$\mathcal{D}$ (best) & -13.0 & 11.3 & 0.439 & 3.53 & 7.12 & 0.61 & 0.88 & -0.19 $\pm$ 1.96\\
\midrule
\multicolumn{9}{c}{\textbf{Constrained Budget $(k=1)$ Oracle Evaluation}}\\
\midrule
$n_{\text{generator}}=\infty$ & -14.6 $\pm$ 0.8 & \textbf{-1.87 $\pm$ 14.9} & 0.439 $\pm$ 0.000 & 3.74 $\pm$ 0.00 & \textbf{6.45 $\pm$ 0.54} & -1.95 $\pm$ 0.00 & 0.88 $\pm$ 0.00 & -0.17 $\pm$	0.29\\
$n_{\text{generator}}=4$ & \textbf{-2.9 $\pm$ 2.2} & -68.6 $\pm$ 109.8 & \textbf{0.571 $\pm$ 0.120} & 3.74 $\pm$ 0.00 & 5.89 $\pm$ 1.42 & -1.95 $\pm$ 0.00 & \textbf{0.89 $\pm$ 0.00} & \textbf{0.01 $\pm$ 0.14}\\
\bottomrule
\toprule
\multicolumn{9}{c}{\textbf{Relaxed Budget $(k=128)$ Oracle Evaluation}}\\
\midrule
$n_{\text{generator}}=\infty$ & -13.3 $\pm$ 0.2 & \textbf{50.2 $\pm$ 2.48} & 0.439 $\pm$ 0.000 & 3.74 $\pm$ 0.00 & 7.38 $\pm$ 0.31 & -1.95 $\pm$ 0.00 & 0.90 $\pm$ 0.01 & \textbf{0.99 $\pm$ 0.01}\\
$n_{\text{generator}}=4$ & \textbf{-1.0 $\pm$ 0.2} & 14.1 $\pm$ 25.0 & \textbf{0.722 $\pm$ 0.091} & 3.74 $\pm$ 0.00 & \textbf{7.98 $\pm$ 0.36} & -1.95 $\pm$ 0.00 & 0.90 $\pm$ 0.01 & 0.95 $\pm$ 0.07\\
\bottomrule
\toprule
\multicolumn{9}{c}{\textbf{Constrained Budget $(k=1)$ Suboptimal (90\%-ile) Oracle Evaluation}}\\
\midrule
$n_{\text{generator}}=\infty$ & -17.0 $\pm$ 1.6 & \textbf{5.88 $\pm$ 4.88} & 0.439 $\pm$ 0.000 & \textbf{3.74 $\pm$ 0.00} & \textbf{7.08 $\pm$ 0.73} & -1.95 $\pm$ 0.00 & 0.89 $\pm$ 0.01 & -1.38 $\pm$ 1.68\\
$n_{\text{generator}}=4$ & \textbf{-14.2 $\pm$ 15.2} & -16.7 $\pm$ 81.1 & \textbf{0.546 $\pm$ 0.148} & 3.22 $\pm$ 0.86 & 6.40 $\pm$ 1.13 & -1.95 $\pm$ 0.00 & 0.89 $\pm$ 0.01 & \textbf{0.24 $\pm$ 0.20}\\
\bottomrule
\end{tabular}}
\end{small}
\end{center}
\end{table*}

\subsection{How does initialization affect the performance of GABO?}

Per \textbf{Algorithm \ref{alg:gabo}}, GABO is based on the BO-qEI baseline optimization policy, which involves initializing the gaussian process (GP) to approximate the offline surrogate model. Consistent with prior work \citep{turbo, maus2022}, we initialize the GP using the pseudo-random Sobol sequence \citep{sobol} at the beginning of the optimization procedure. However, an alternative approach is to instead initialize the GP using the top $n_{\text{init}}$ samples from the offline dataset. In particular, this strategy is already employed in both related work describing the baseline first-order optimization methods assessed herein, with the idea that better designs can be generated by initializing from better designs. We compare these two GP initialization strategies in \textbf{Supplementary Table \ref{table:ablate-gp-initialization}}.

Interestingly, our results show that initializing the GABO GP from the Sobol sequence consistently outperforms initialization from the top candidates in offline dataset. We hypothesize that this may be due to the fact that top-scoring candidates likely lie in similar regions of the input space, which significantly alters the ability of the optimizer to explore other regions of the design space over the course of the optimization process. Future work may help better interrogate the relationship between GP initialization and offline optimization, which is outside the scope of this work.

\begin{table*}[t]
\caption{\textbf{GABO GP Initialization Ablation Study}$\quad$ We investigate the effect of initializing the Gaussian process (GP) in GABO using the best $n_{\text{init}}$ points from the offline dataset (i.e., \textbf{Best} initialization strategy) versus our method in \textbf{Algorithm \ref{alg:gabo}} where the GP is initialized using the first $n_{\text{init}}$ points from the Sobol sequence from \citep{sobol} (i.e., \textbf{Sobol} initialization strategy). Oracle values are averaged across 10 random seeds and reported as mean $\pm$ standard deviation. In each evaluation setting, we rank all 2,048 proposed designs according to the penalized surrogate forward model in (\ref{eq:lagrangian}) and evaluate the top $k$ designs using the oracle function, reporting the maximum out of the $k$ oracle values. In the suboptimal evaluation setting, we report the oracle score of the single 90th percentile design according to the penalized surrogate ranking. \textbf{Bold} entries indicate the best entry in the column for the particular optimizer and evaluation metric. $^*$Denotes the life sciences MBO tasks offered by Design-Bench \citep{design-bench}.\vspace{-2ex}}
\label{table:ablate-gp-initialization}
\begin{center}
\begin{small}
\resizebox{\textwidth}{!}{\begin{tabular}{rccccccccc}
\toprule
\textbf{Strategy} & \textbf{Branin} & \textbf{LogP} & \textbf{TF-Bind-8$^*$} & \textbf{GFP$^*$} & \textbf{UTR$^*$} & \textbf{ChEMBL$^*$} & \textbf{D'Kitty} & \textbf{Warfarin}\\
\midrule
$\mathcal{D}$ (best) & -13.0 & 11.3 & 0.439 & 3.53 & 7.12 & 0.61 & 0.88 & -0.19 $\pm$ 1.96\\
\midrule
\multicolumn{9}{c}{\textbf{Constrained Budget $(k=1)$ Oracle Evaluation}}\\
\midrule
Best & -3.6 $\pm$ 4.1 & 14.0 $\pm$ 18.4 & 0.504 $\pm$ 0.117 & 2.97 $\pm$ 1.02 & 5.36 $\pm$ 1.24 & \textbf{0.61 $\pm$ 0.00} & 0.50 $\pm$ 0.19 & -2.97 $\pm$ 9.03\\
Sobol & \textbf{-2.6 $\pm$ 1.1} & \textbf{21.3 $\pm$ 33.2} & \textbf{0.570 $\pm$ 0.131} & \textbf{3.60 $\pm$ 0.40} & \textbf{7.51 $\pm$ 0.39} & 0.60 $\pm$ 0.07 & \textbf{0.71 $\pm$ 0.01} & \textbf{0.60 $\pm$ 1.80}\\
\bottomrule
\toprule
\multicolumn{9}{c}{\textbf{Relaxed Budget $(k=128)$ Oracle Evaluation}}\\
\midrule
Best & -0.5 $\pm$ 0.0 & 118.9 $\pm$ 19.5 & 0.918 $\pm$ 0.034 & 3.74 $\pm$ 0.00 & \textbf{8.37 $\pm$ 0.09} & 0.66 $\pm$ 0.01 & \textbf{0.87 $\pm$ 0.05} & 0.99 $\pm$ 0.09\\
Sobol & -0.5 $\pm$ 0.1 & \textbf{122.1 $\pm$ 20.6} & \textbf{0.954 $\pm$ 0.025} & 3.74 $\pm$ 0.00 & 8.36 $\pm$ 0.08 & \textbf{0.70 $\pm$ 0.01} & 0.72 $\pm$ 0.00 & \textbf{1.00 $\pm$ 0.03} \\
\bottomrule
\toprule
\multicolumn{9}{c}{\textbf{Constrained Budget $(k=1)$ Suboptimal (90\%-ile) Oracle Evaluation}}\\
\midrule
Best & \textbf{-11.8 $\pm$ 6.4} & -85.9 $\pm$ 124 & 0.382 $\pm$ 0.106 & 3.45 $\pm$ 0.77 & \textbf{6.28 $\pm$ 1.70} & 0.60 $\pm$ 0.03 & \textbf{0.64 $\pm$ 0.23} & -0.65 $\pm$ 3.97\\
Sobol & -12.7 $\pm$ 10.0 & \textbf{-12.2 $\pm$ 46.1} & \textbf{0.467 $\pm$ 0.066} & \textbf{3.56 $\pm$ 1.66} & 6.12 $\pm$ 1.22 & \textbf{0.61 $\pm$ 0.08} & 0.57 $\pm$ 0.17 & \textbf{0.02 $\pm$ 5.77}\\
\bottomrule
\end{tabular}}
\end{small}
\end{center}
\end{table*}

\subsection{Can the Gaussian process (GP) in GABO be directly used as the surrogate forward model?}

In \textbf{Algorithm \ref{alg:gabo}}, we leverage a surrogate forward model $f_{\theta}$ in model-based optimization and a separate GP to acquire samples in the Bayesian optimization framework. However, it may be possible to use the GP directly as the surrogate forward model. Our results in \textbf{Supplementary Table \ref{table:ablate-surrogate}} suggest that this is \textit{not} an effective strategy with which to use GABO\textemdash using even the simple neural-network as the surrogate function (as done in our approach in \textbf{Algorithm \ref{alg:gabo}}) outperforms the alternative GP-based approach in six of the eight tasks in the top-1 evaluation setting, and is non-inferior to the alternative GP-based approach in all eight tasks in the top-128 evaluation setting. These results suggest that using a more complex neural-network surrogate function for GABO leads to better optimization results than directly using the GP as the surrogate function.

\begin{table*}[t]
\caption{\textbf{GABO Neural Network Surrogate Ablation Study}$\quad$ Instead of using a neural network (NN) as our surrogate forward model, we explore if the Gaussian process (GP) employed by the parent BO optimizer can directly be used as the surrogate model in GABO's framwork. Oracle values are averaged across 10 random seeds and reported as mean $\pm$ standard deviation. In each evaluation setting, we rank all 2,048 proposed designs according to the penalized surrogate forward model in (\ref{eq:lagrangian}) and evaluate the top $k$ designs using the oracle function, reporting the maximum out of the $k$ oracle values. In the suboptimal evaluation setting, we report the oracle score of the single 90th percentile design according to the penalized surrogate ranking. \textbf{Bold} entries indicate the best entry in the column for the particular optimizer and evaluation metric. $^*$Denotes the life sciences MBO tasks offered by Design-Bench \citep{design-bench}.\vspace{-2ex}}
\label{table:ablate-surrogate}
\begin{center}
\begin{small}
\resizebox{\textwidth}{!}{\begin{tabular}{rccccccccc}
\toprule
\textbf{Surrogate} & \textbf{Branin} & \textbf{LogP} & \textbf{TF-Bind-8$^*$} & \textbf{GFP$^*$} & \textbf{UTR$^*$} & \textbf{ChEMBL$^*$} & \textbf{D'Kitty} & \textbf{Warfarin}\\
\midrule
$\mathcal{D}$ (best) & -13.0 & 11.3 & 0.439 & 3.53 & 7.12 & 0.61 & 0.88 & -0.19 $\pm$ 1.96\\
\midrule
\multicolumn{9}{c}{\textbf{Constrained Budget $(k=1)$ Oracle Evaluation}}\\
\midrule
GP & -37.4 $\pm$ 4.4 & -57.9 $\pm$ 159.2 & \textbf{0.576 $\pm$ 0.058} & 3.51 $\pm$ 0.69 & 6.84 $\pm$ 1.24 & \textbf{0.65 $\pm$ 0.01} & 0.42 $\pm$ 0.17 & -0.28 $\pm$ 2.13 \\
NN & \textbf{-2.6 $\pm$ 1.1} & \textbf{21.3 $\pm$ 33.2} & 0.570 $\pm$ 0.131 & \textbf{3.60 $\pm$ 0.40} & \textbf{7.51 $\pm$ 0.39} & 0.60 $\pm$ 0.07 & \textbf{0.71 $\pm$ 0.01} & \textbf{0.60 $\pm$ 1.80}\\
\bottomrule
\toprule
\multicolumn{9}{c}{\textbf{Relaxed Budget $(k=128)$ Oracle Evaluation}}\\
\midrule
GP & -1.5 $\pm$ 0.5 & 119.9 $\pm$ 20.1 & 0.755 $\pm$ 0.071 & 3.74 $\pm$ 0.00 & 8.34 $\pm$ 0.07 & 0.67 $\pm$ 0.01 & 0.72 $\pm$ 0.00 & -0.27 $\pm$ 2.13 \\
NN & \textbf{-0.5 $\pm$ 0.1} & \textbf{122.1 $\pm$ 20.6} & \textbf{0.954 $\pm$ 0.025} & 3.74 $\pm$ 0.00 & \textbf{8.36 $\pm$ 0.08} & \textbf{0.70 $\pm$ 0.01} & 0.72 $\pm$ 0.00 & \textbf{1.00 $\pm$ 0.03} \\
\bottomrule
\toprule
\multicolumn{9}{c}{\textbf{Constrained Budget $(k=1)$ Suboptimal (90\%-ile) Oracle Evaluation}}\\
\midrule
GP & \textbf{-10.1 $\pm$ 10.6} & -51.5 $\pm$ 108.8 & \textbf{0.562 $\pm$ 0.091} & 2.62 $\pm$ 1.13 & \textbf{6.54 $\pm$ 1.56} & \textbf{0.65 $\pm$ 0.00} & 0.50 $\pm$ 0.19 & -0.27 $\pm$ 2.13 \\
NN & -12.7 $\pm$ 10.0 & \textbf{-12.2 $\pm$ 46.1} & 0.467 $\pm$ 0.066 & \textbf{3.56 $\pm$ 1.66} & 6.12 $\pm$ 1.22 & 0.61 $\pm$ 0.08 & \textbf{0.57 $\pm$ 0.17} & \textbf{0.02 $\pm$ 5.77}\\
\bottomrule
\end{tabular}}
\end{small}
\end{center}
\end{table*}

\subsection{What is the computational cost of running aSCR (i.e., \textbf{Algorithm \ref{alg:ascr}})?}

At first glance, Adaptive SCR may appear to be a computationally expensive algorithm: it requires us to dynamically re-train a source critic neural network and compute the Lagrangian hyperparameter at each step through a grid search. However, in the implementation used for our experiments, the grid search to compute $\alpha$ is highly vectorized, and the source critic re-training patience and learning rate are such that the computational cost from re-training is not too significant. As a result, we are able to run Adaptive SCR with both Bayesian Optimization (BO) and Gradient Ascent (GA) using an experimental setup with one 24-core Intel Xeon CPU and one NVIDIA RTX A6000 GPU. To benchmark our implementation, we evaluate BO and GA both with and without our Generative Adversarial (GA) source critic regularization algorithm on the \textbf{Branin} and Penalized \textbf{LogP} optimization tasks. As a reminder, the Branin task is a standard benchmarking task for offline optimization, and the Penalized LogP task is subjectively the most challenging task assessed in our manuscript with the highest dimensional design space out of the eight assessed tasks.

Our results are shown in \textbf{Supplementary Table \ref{table:timing}}. On the Branin toy task, aSCR increases the compute time by 257\% for BO and 680\% for GA, which is a significant computational cost. However, on the more challenging \textbf{LogP} task more representative of the tasks encountered in the applications of offline optimization, aSCR only introduces a 6.9\% increase in compute time for GA and 28.9\% increase for BO. Furthermore, while there are evidently additional compute costs associated with running our aSCR algorithm, we note that in most applications of offline optimization, obtaining labeled data is the main bottleneck in many practical applications. Thus, it is often worth spending this extra compute to ensure the best results for a given evaluation budget using aSCR.

\begin{table*}[t]
\caption{\textbf{Computational Tractability}$\quad$ Runtimes on a single node using one NVIDIA RTX A6000 GPU are averaged across 10 random seeds and reported as mean $\pm$ standard deviation.}
\label{table:timing}
\begin{center}
\begin{small}
\resizebox{0.8\textwidth}{!}{\begin{tabular}{rccccrcc}
\toprule
\textbf{Time (sec)} & \textbf{Branin} & \textbf{LogP} & & & \textbf{Time (sec)} & \textbf{Branin} & \textbf{LogP}\\
\midrule
Grad. & 9.68 $\pm$ 0.23 & 765 $\pm$ 6.64 & & &  BO & 92.1 $\pm$ 10.2 & 965 $\pm$ 16.8 \\
GAGA & 75.6 $\pm$ 25.4 & 818 $\pm$ 10.5 & & & GABO & 329 $\pm$ 146 & 1245 $\pm$ 55.2 \\
\midrule
\% Increase & 680\% $\pm$ 259\% & 6.9\% $\pm$ 1.6\% & & & \% Increase & 257\% $\pm$ 157\% & 28.9\% $\pm$ 4.5\% \\
\bottomrule
\end{tabular}}
\end{small}
\end{center}
\end{table*}

\subsection{How do the performance of GABO and other optimization methods vary with the allowed oracle query budget \texorpdfstring{$k$}{k}?}
\begin{figure}[htbp]
    \centering
    \includegraphics[width=\textwidth]{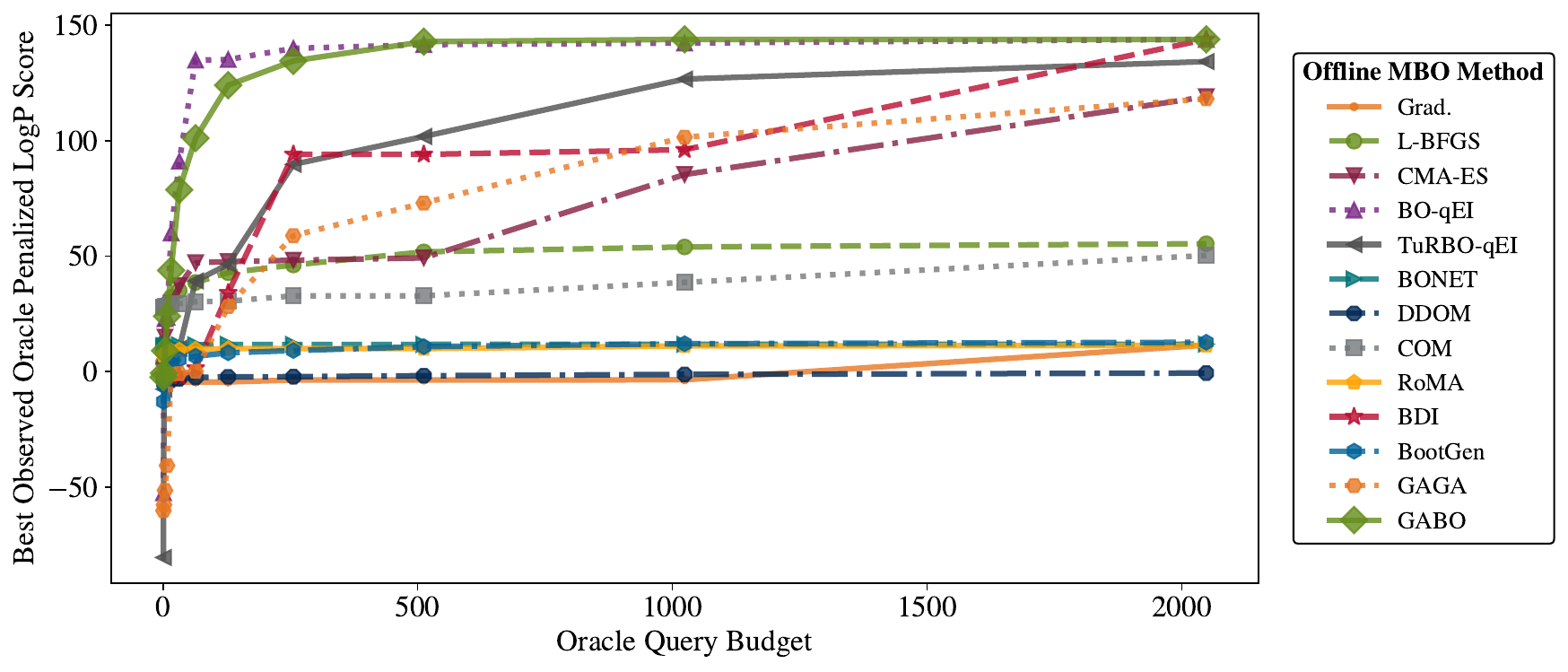}
    \vskip -0.1in
    \caption{\textbf{100th Percentile Oracle Scores versus $k$-Shot Oracle Budget Size}$\quad$ We plot the 100th percentile oracle Penalized LogP score averaged across 10 random seeds as a function of the number of allowed oracle calls $k$.}
    \label{fig:budget}
\end{figure}

To investigate this question, we vary the number of allowed $k$-shot oracle calls in the Penalized \textbf{LogP} task (\textbf{Supplementary Fig. \ref{fig:budget}}). While the majority of first-order optimization methods we evaluated are able to reach local optima rapidly, the proposed designs from such approaches are suboptimal compared to those from GABO (and GAGA) with Adaptive SCR as the oracle query budget size increases. Separately, comparing the curves for GABO and vanilla BO-qEI, we see that GABO with Adaptive SCR is able to propose consistently superior design candidates in the small query budget regime often encountered in real-world settings. This is due to the fact that GABO regularizes the surrogate function estimates such that the proposed candidates are both high-scoring according to the surrogate objective \textit{and} relatively in-distribution. Our results demonstrate that especially for real-world tasks like molecule design with complex objective function landscapes, methods such as GABO with Adaptive SCR are able to explore diverse, high-performing design candidates effectively even in the setting of small oracle query budgets.

\subsection{Is the optimization budget sufficient for optimization convergence?}
\begin{figure}
    \centering
    \includegraphics[width=\textwidth]{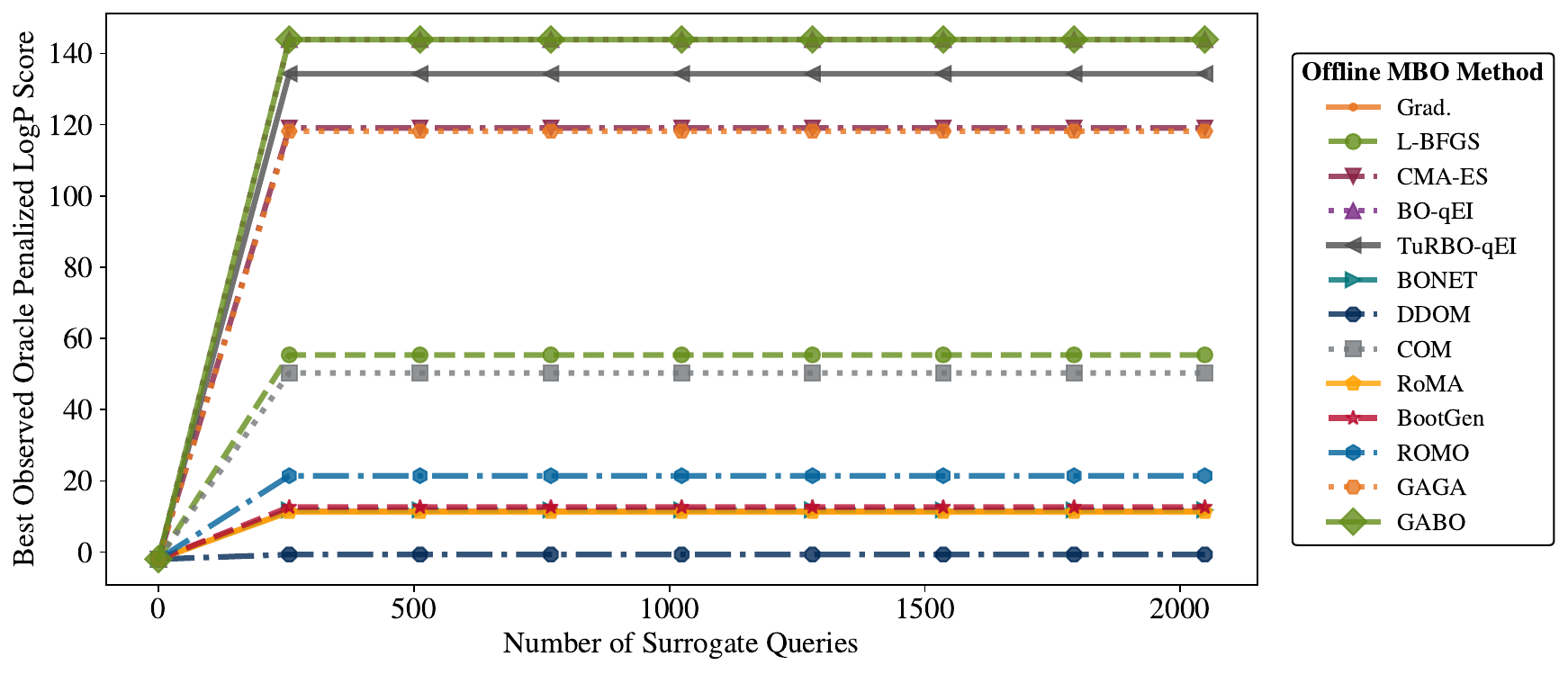}
    \caption{\textbf{Best Oracle Penalized LogP Value versus Optimization Step Count}\quad We plot the best Penalized LogP score averaged across 10 random seeds as a function of the number of surrogate queries made over the optimization trajectory. All offline model-based optimization (MBO) methods assessed consistently converge within the allowed oracle query budget used in our experimental setup as described in \textbf{Section \ref{section:experimental}}.}
    \label{fig:convergence}
    \vspace{5.78in}
\end{figure}

For all of our experimental results, we restrict the surrogate query budget to a total of 2048 allowed offline surrogate model queries in order to ensure a fair comparison between different optimization methods. To ensure that such a budget is sufficient for optimizer convergence across different optimization methods, we plot the best achieved oracle Penalized LogP value (i.e., assuming an unlimited oracle evaluation budget) as a function of the number of optimizer surrogate queries (\textbf{Supplementary Fig. \ref{fig:convergence}}) for the Penalized LogP task. These results show that our methods are indeed able to converge over the course of the optimization trajectory.

\end{document}